\documentclass[a4paper,journal]{IEEEtran}
\usepackage{color,soul}
\usepackage{amsmath}
\usepackage{amssymb}
\usepackage{cite}
\usepackage{amsfonts}
\usepackage{graphicx}
\usepackage{subfig}
\usepackage{tabularray}
\usepackage{bbm}
\usepackage[table]{xcolor}
\usepackage{balance}
\usepackage{multirow}
\usepackage[table]{xcolor}
\definecolor{lavender}{rgb}{0.9, 0.9, 0.98}
\usepackage[linesnumbered,ruled,vlined]{algorithm2e}
\usepackage{caption}
\usepackage{comment}
\usepackage{booktabs}
\usepackage{dirtytalk}
\usepackage{soul}
\usepackage{booktabs}
\usepackage{xr}

\newtheorem{definition}{Definition}

\SetKwInput{KwInput}{Input}                % Set the Input
\SetKwInput{KwOutput}{Output}

\newcommand{\sN}{\mathcal{N}}
\newcommand{\setC}{\mathbb{C}}
\newcommand{\sC}{\mathcal{C}}
\newcommand{\bX}{\boldsymbol{X}}
\newcommand{\bY}{\boldsymbol{Y}}
\newcommand{\bC}{\boldsymbol{C}}

\newcommand{\bV}{\boldsymbol{V}}
\newcommand{\bS}{\boldsymbol{S}}

\newcommand{\bL}{\boldsymbol{L}}
\newcommand{\GOAI}{\mathtt{GOAI}}
\ifCLASSINFOpdf
  
\else
 
\fi

\begin{document}
\title{\huge SC-GIR: Goal-oriented Semantic Communication via Invariant Representation Learning for Image Transmission}
\author{Senura Hansaja Wanasekara, Van-Dinh Nguyen, Kok-Seng Wong,
         \\ M.-Duong Nguyen, Symeon Chatzinotas, and Octavia~A.~Dobre
\thanks{S. H. Wanasekara is with the University of Sydney, Sydney, Australia, and was with the College of Engineering and Computer Science, VinUniversity, Hanoi, Vietnam  (e-mail: wwan0281@uni.sydney.edu.au).}
\thanks{K.-S. Wong is with the College of Engineering and Computer Science, and also with the Center for Environmental Intelligence (CEI), VinUniversity, Hanoi 100000, Vietnam  (e-mail: wong.ks@vinuni.edu.vn). }
\thanks{V.-D. Nguyen is wih the School of Computer Science and Statistics, Trinity College Dublin, Dublin 2 D02PN40, Ireland, and was with the College of Engineering and Computer Science, and also with the Center for Environmental Intelligence (CEI), VinUniversity, Hanoi 100000, Vietnam  (e-mail: dinh.nguyen@tcd.ie). Corresponding author: 
\textit{Van-Dinh Nguyen}.}
\thanks{M.-D. Nguyen is with the Department of Intelligent Computing and Data Science, VinUniversity, Hanoi, Vietnam (e-mail: duong.nm2@vinuni.edu.vn).}
\thanks{S. Chatzinotas is with the Interdisciplinary Centre for Security, Reliability and Trust (SnT), University of Luxembourg, L-1855 Luxembourg City, Luxembourg (e-mail: symeon.chatzinotas@uni.lu).}
\thanks{O. A. Dobre is with the Dept. of Electrical and Computer Engineering, Memorial University, St. John’s, NL A1B 3X9, Canada (e-mail: odobre@mun.ca).}
\thanks{This work was supported in part by CEI under Grant VUNI.CEI.FS\_0002 and the Luxembourg National Research Fund, grant reference C23/IS/18073708/SENTRY. The work of V.-D. Nguyen was supported by the Australia-Vietnam Strategic Technologies Centre (AVSTC) Seed Grant. Part of this work was presented at the 11th EAI International Conference on Industrial Networks and Intelligent Systems (EAI INISCOM), Da Nang, Vietnam, Feb. 2025.}
}

\markboth{IEEE Transactions on Mobile Computing}%
{Shell \MakeLowercase{\textit{et al.}}: SC-GIR: Goal-oriented Semantic Communication via Invariant Representation Learning}
\maketitle

\begin{abstract}
Goal-oriented semantic communication (SC) aims to revolutionize communication systems by transmitting only task-essential information. However, current approaches face challenges such as joint training at transceivers, leading to redundant data exchange and reliance on labeled datasets, which limits their task-agnostic utility.
To address these challenges, we propose a novel framework called Goal-oriented Invariant Representation-based SC (SC-GIR) for image transmission. Our framework leverages self-supervised learning to extract an invariant representation that encapsulates crucial information from the source data, independent of the specific downstream task. This compressed representation facilitates efficient communication while retaining key features for successful downstream task execution. Focusing on machine-to-machine tasks, we utilize covariance-based contrastive learning techniques to obtain a latent representation that is both meaningful and semantically dense. To evaluate the effectiveness of the proposed scheme on downstream tasks, we apply it to various image datasets for lossy compression. The compressed representations are then used in a goal-oriented AI task. Extensive experiments on several datasets demonstrate that SC-GIR outperforms baseline schemes by nearly $10\%$, and achieves over $85\%$ classification accuracy for compressed data under different SNR conditions. These results underscore the effectiveness of the proposed framework in learning compact and informative latent representations.
\end{abstract}


\IEEEpeerreviewmaketitle

\section{Introduction}
{\IEEEPARstart{T}{he} rapid evolution of wireless communication technologies, particularly with the widespread deployment of the fifth-generation of cellular network technology (5G), has catalyzed a new wave of intelligent and data-intensive applications, such as digital twins, smart cities, and the growing ecosystem of the Internet of Things (IoT)~\cite{2024-SemCom-Survey2}. However, conventional communication paradigms remain largely centered on the transmission of raw data, paying little attention to the semantic content or contextual intent underlying the information being exchanged. This traditional framework, firmly grounded in Shannon's classical information theory, is increasingly showing its limitations in the face of modern demands for intelligent, purpose-driven, and resource-efficient communication.

As we look ahead to the sixth-generation (6G) wireless networks, the landscape of communication is poised to undergo a transformative shift. Next-generation applications, ranging from immersive technologies such as virtual and augmented reality (VR/AR) to the Metaverse and real-time digital twinning, require unprecedented data rates and ultra-low latency. Many of these applications push the boundaries of Shannon’s capacity limits~\cite{cabrera20216g}, while simultaneously introducing new requirements such as contextual understanding, semantic fidelity, and task relevance.

Moreover, the convergence of artificial intelligence (AI) with wireless communications is steering the field toward user-centric and context-aware paradigms. In these settings, the meaning and intent of transmitted information often outweigh the need to reproduce data with bit-level accuracy. This trend emphasizes that future communication systems must evolve from merely delivering bits to delivering purpose-driven understanding \cite{ZhijinProc24}.

Semantic communication (SC) has emerged as a promising response to these challenges. By prioritizing the meaning embedded in data rather than its raw quantity, SC integrates AI-based approaches, representation learning, and context awareness to enhance the efficiency and effectiveness of information exchange \cite{ZhijinProc24,9994683}. This paradigm shift aims to significantly reduce bandwidth consumption, increase task-specific reliability, and improve the end-user experience. As such, developing robust semantic communication frameworks is essential to overcoming the bottlenecks of traditional communication models and unlocking the full potential of 6G and beyond.

\subsection{Challenges}
Despite its promise, the practical adoption of SC, especially in distributed and resource-constrained environments such as IoT networks, faces several significant challenges. Most existing SC systems rely on end-to-end autoencoder architectures that aim to fully reconstruct transmitted data at the receiver. While effective for certain tasks, this approach falls short in scenarios that prioritize goal-oriented outcomes over pixel-accurate reconstructions. In goal-oriented SC, full data reconstruction introduces unnecessary redundancy that does not contribute to the intended downstream task (\textit{e.g.} classification or detection). This not only wastes bandwidth but also increases computational and energy costs at the receiver side \cite{hu2022robust, shao2021learning}. In federated and collaborative learning scenarios, reconstructing high-fidelity data at distributed nodes can inadvertently expose sensitive information, raising critical privacy and security concerns~\cite{10458014, 10483549}. Moreover, achieving near-perfect reconstruction often necessitates massive labeled training datasets, which are difficult to acquire and maintain in dynamic and heterogeneous IoT environments~\cite{9252948}.

Traditional wireless communication systems, such as those based on the joint source-channel coding (JSCC)~\cite{bourtsoulatze2019deep}, transmit images by first compressing them into latent representations $\boldsymbol{x} \in \mathbb{R}^n$ using an encoder $f_{\boldsymbol{\theta}}$, sending these representations over a channel $h$, and decoding them at the receiver using a decoder $g_{\boldsymbol{\phi}}$. While JSCC has demonstrated robustness to channel noise, it faces notable limitations when applied in IoT and edge computing contexts:
\begin{itemize}
    \item \textbf{Scalability challenges}: DeepJSCC models require retraining or fine-tuning for new environments or device configurations, limiting their adaptability in diverse and rapidly changing IoT deployments.
    \item \textbf{Redundant reconstruction}: The decoder often upscales latent features into high-dimensional data (\textit{e.g.} full-resolution images), much of which may be irrelevant for the actual task. This not only introduces computational inefficiencies but can also dilute the semantic signal with unnecessary noise.

    \item \textbf{Encoding and power constraints}: Transmitting latent vectors $\boldsymbol{z}$ composed of $k$-length elements under strict power and bandwidth constraints complicates system design and increases implementation complexity, especially in edge scenarios with limited energy budgets.
\end{itemize}
}

\subsection{Main Contributions}
Motivated by the limitations of existing SC paradigms, we propose a novel framework, \textit{namely} SC-GIR, tailored for efficient and goal-oriented image transmission in wireless environments. SC-GIR is grounded in the principles of the information bottleneck theory and enhanced by advanced contrastive learning techniques~\cite{liu2021self}, enabling it to selectively extract and transmit semantically relevant information while discarding irrelevant details. This design effectively addresses the challenges of redundancy, privacy, and scalability in semantic communication for IoT scenarios. The key contributions of this paper are summarized as follows:

\begin{enumerate}
\item We propose {SC-GIR}, a novel semantic communication framework for image transmission, explicitly designed to reduce transmission overhead in goal-oriented tasks. By prioritizing semantically meaningful representations over pixel-level fidelity, SC-GIR significantly improves communication efficiency in resource-constrained environments.

\item We develop a specialized training methodology for SC-GIR that leverages self-supervised contrastive learning to minimize semantic redundancy. This approach ensures that only task-relevant features are retained, thereby enhancing both communication efficiency and task performance, without the need for full data reconstruction.

\item We conduct extensive experiments across multiple benchmark datasets to validate SC-GIR in the context of IoT-enabled semantic classification tasks. Our results demonstrate that SC-GIR achieves substantial improvements in both transmission efficiency and task accuracy, while also exhibiting superior scalability and robustness compared to state-of-the-art semantic communication baselines.
\end{enumerate}

\subsection{Organization and Notation}
The remainder of this paper is organized as follows.  The related works are discussed in Section \ref{sec_Related}. Section \ref{sec_Model} presents the system model along with the objective design.  The proposed SC-GIR framework is detailed in Section \ref{sec_Proposed}. Numerical results are given in Section \ref{sec_Numerical}, while  Section \ref{sec_Conclusion} concludes the paper.

\textit{Notation:} Throughout this paper, we use the following notations described in Table \ref{tab:notations}. 
\begin{table}[!hbt]{
\caption{Mathematical Notations}
\label{tab:notations}
\centering
\begin{tabular}{c l}
\toprule
\textbf{Symbol} & \textbf{Description} \\
\cmidrule(lr){1-1}  \cmidrule(lr){2-2}
\textbf{A} & Matrix \\
\textbf{a} & Vector \\
a & Scalar \\
$x \sim \sC\sN(0,\sigma^2)$ & \begin{tabular}[c]{@{}l@{}}Circularly-symmetric complex Gaussian\\  random variable with zero-mean and
variance $\sigma^2$\end{tabular} \\
$\|\cdot\|$ & Euclidean norm of a vector \\
$\mathbbm{1}_{\{\}}$ & Indicator function \\
$\setC$ & Set of complex numbers \\
$\mathbb{R}$ & 	Set of real numbers \\
$\mathbb{E}\{\cdot\}$ & \begin{tabular}[c]{@{}l@{}}Expectation of a random variable  variable\end{tabular} \\
$\mathcal{L}\{\cdot\}$ & Loss function \\
$\mathcal{C}_{ij}$ & Cross-correlation matrix element \\
$I(\mathbf{X};\mathbf{S})$ & Mutual information between $\mathbf{X}$ and $\mathbf{S}$\\
\bottomrule
\end{tabular}}
\end{table}

\section{Related Works}\label{sec_Related}
In communication systems, Shannon's foundational model emphasizes the conversion of information into bit sequences for transmission, prioritizing accuracy as measured by metrics such as bit error rate (BER) and symbol error rate (SER) \cite{shannon1948mathematical}. While effective for ensuring data fidelity, this traditional approach often neglects the semantic value of transmitted information, a crucial aspect for understanding and utilizing data effectively \cite{jerri1977shannon}. With the evolution of digital communication, network protocols have become increasingly complex, posing challenges in compatibility and increased network complexity \cite{shi2021semantic, al2017internet, bayilmics2022survey}. In response, the concept of semantic communication has emerged, advocating for a shift towards transmitting the essence or meaning of data, rather than just the data itself. This paradigm shift is particularly relevant in machine-to-machine communications, where the efficiency and relevance of information exchange are of paramount importance \cite{stankovic2014research}.

Existing research in SC can broadly be categorized based on its encoder-decoder design strategies, particularly regarding how they represent, compress, and interpret semantic information. One prominent direction focuses on improving semantic encoding and decoding efficiency, typically by integrating deep learning techniques to balance compression and task performance. DeepSC~\cite{2021-SEM-DeepSC}, a transformer-based architecture, was one of the earliest efforts to incorporate natural language understanding into SC systems. Its successors, U-DeepSC~\cite{2023-SemCom-UDeepSC} and MUDeepSC~\cite{2022-SemCom-MUDeepSC}, extend this model by leveraging layer-wise knowledge transfer to improve generalization across tasks. Mem-DeepSC~\cite{2023-SemCom-MemDeepSC} further enhances this architecture with memory modules at the receiver side, enabling better contextual understanding and reconstruction of semantic content. Another influential line of work is DeepJSCC~\cite{bourtsoulatze2019deep}, which eliminates explicit source and channel coding stages by jointly mapping input images directly to channel symbols using deep neural networks. While highly robust in noisy wireless conditions, DeepJSCC suffers from scalability issues and task-specific limitations, especially in heterogeneous or dynamic IoT environments.

Recent works have addressed the adaptability and generalization challenges of SC systems. For example, adaptable semantic compression techniques~\cite{2022-SEM-AdaptableSemanticCompression, 2020-SemCom-DJSCCF, 2019-SemCOm-DJSCC-WIT, 2024-SemCom-AdaSem} dynamically assessed the relevance of latent features to discard information that contributes little to task performance. This approach significantly reduces bandwidth while maintaining high accuracy in downstream tasks.
SemCC~\cite{2024-SemCom-SemCC} introduced contrastive learning into the encoding process, allowing the model to extract more salient features that are robust to variations in input. DeepMA~\cite{2024-SemCom-DeepMA} proposed orthogonal semantic symbol extraction, supporting more efficient multiple access schemes by separating semantic vectors across users.
A different yet promising direction is explored in GenerativeJSCC~\cite{2023-SemCom-GenerativeJSCC}, which integrates StyleGAN-2 to leverage pre-trained generative priors. This approach offers high-quality reconstructions by aligning transmitted representations with generative latent spaces. However, its reliance on domain-specific pretraining introduces limitations in generalizability to new or unseen datasets, a critical issue for real-world IoT and edge deployments.

Beyond methods explicitly designed for semantic communication, two adjacent domains offer relevant insights into task-agnostic and compression-efficient modeling. The first involves large-scale, multi-modal models like CLIP (Contrastive Language-Image Pre-training) \cite{radford21a}, which learn highly generalizable, task-agnostic representations from vast image-text datasets, enabling remarkable zero-shot transfer capabilities. However, these models differ from M2M communications in their objectives and constraints; their primary goal is multi-modal understanding and their massive size makes them largely unsuitable for the resource-constrained edge devices central to our work. The second relevant domain is that of generative models like Variational Autoencoders (VAEs) \cite{lopez2018information}, which are inherently designed for data compression. While some VAEs incorporate semantic regularization, their fundamental objective remains data reconstruction. M2M communication is distinct from both approaches. Unlike large-scale models, M2M communication operates under strict constraints, prioritizing lightweight and communication-efficient architectures.

Despite these advances, most existing methods still operate under the assumption that full or near-full reconstruction of the original input is necessary. This contradicts the objective of many goal-oriented applications, where only task-relevant semantic features are required. Moreover, these frameworks often neglect concerns such as privacy, redundancy, and scalability in distributed environments. This creates an urgent need for frameworks that not only reduce bandwidth usage but also preserve utility, generalize across tasks, and ensure data privacy, particularly in decentralized or federated settings.

\section{System Model}\label{sec_Model}
\begin{figure}[t]
    \centering
    \includegraphics[width=0.5\textwidth]{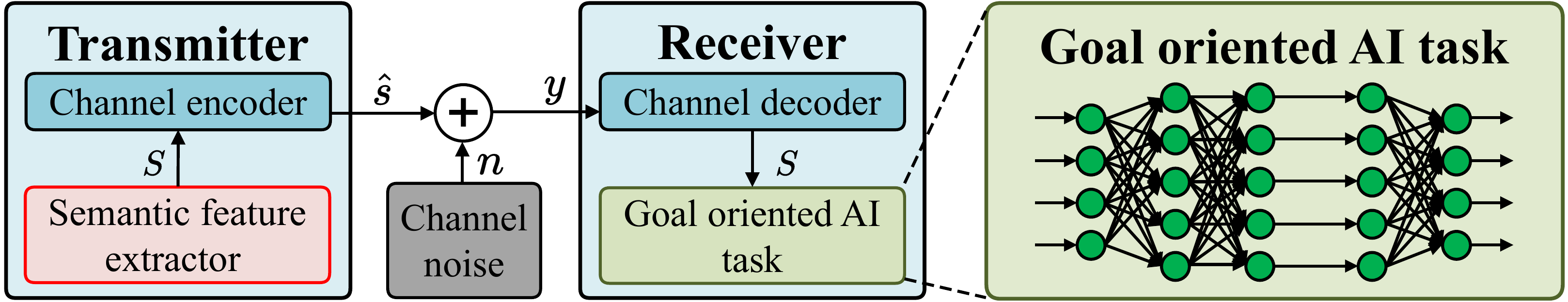}
    \caption{The overall architecture of the communication framework.}
    \label{fig:system_model}
\end{figure}

As illustrated in Fig.~\ref{fig:system_model}, the overall architecture of the proposed SC-GIR framework targets a point-to-point communication scenario with a transmitter and receiver. The transceiver utilizes the latent representation for downstream AI tasks such as classification or object detection. The data undergoes processing through key system components highlighted in different colors. We will delve into each component's functionality in subsequent sections. To contextualize our discussion, we consider a harvested data sample denoted by $\boldsymbol{x} \in \mathbb{R}^{c \times d\times w}$, where $c$, $d$ and $w$ represent the channels, height and width, respectively, encapsulating the dimensional attributes of the data pertinent to tasks.

\subsection{Wireless Channel}
The transmission of the channel-encoded latent variable $\hat{\boldsymbol{s}}$, obtained via the channel encoder,  occurs over a Rayleigh fading channel, a common scenario in wireless communications characterized by multipath scattering effects \cite{xie2021deep}. We assume that ${\hat{\boldsymbol{s}}}\in\mathbb{C}^{N\times 1}$, where $N$ is the output dimension of the semantic encoder. The received signal can be expressed as
\begin{align}\label{eq:wirelesschannel}
    {\boldsymbol{y}} = h{\hat{\boldsymbol{s}}} + \boldsymbol{n}
\end{align}
where $h\sim\mathcal{CN}(0,1)$ is the channel coefficient characteristic of Rayleigh fading and $\boldsymbol{n}\sim\mathcal{CN}(0,\sigma_n^2)$ is the additive white Gaussian noise (AWGN). Eq. \eqref{eq:wirelesschannel} captures the essential aspects of signal degradation over a wireless channel, highlighting the influence of both multipath fading and thermal noise in the received signal ${\boldsymbol{y}}$.

\subsection{Semantic Feature Extractor} The proposed semantic feature extractor plays a critical role in achieving a concise and informative latent representation of the data. This extractor acts as the foundation for understanding the underlying meaning of the data. The proposed algorithm presented shortly operates on the data $\boldsymbol{X}$, analyzing and distilling complex information into meaningful features represented by the latent representation $\bS$ (in here we consider $\bS \sim \bS'$ for explanation purposes). {The work in \cite{federici2020learning} proved that a representation $\bS$ is minimally sufficient for downstream tasks $\bY$ if and only if $I(\bX;\bY|\bS) = 0 $ (once $\bS$ is known, $\bX$ provides no additional information about $\bY$)}. This implies that the encoding process preserves all task-relevant information. To achieve this, we aim to minimize the following loss function:
\begin{align}\label{eq:information_loss}
    \mathcal{L} = I(\bX;\bS) - \beta I(\bS;\bY).
\end{align}
The loss function consists of two key information-theoretic terms: $I(\bX;\bS)$ and $I(\bS;\bY)$. In particular, $I(\bX;\bS)$ represents the mutual information between the original data $\bX$ and the latent representation $\bS$. It captures how much information about the original data is preserved in the encoded representation. $I(\bS;\bY)$ presents the mutual information between the latent representation $\bS$ and downstream task $\bY$ weighted by a hyper-parameter $\beta$. It controls the trade-off between preserving task-relevant information and achieving compression.

By minimizing the loss function in Eq.~\eqref{eq:information_loss}, we aim to learn latent representations that are informative for the downstream task while also achieving compression by removing redundant information about the original data. To achieve this, we decompose $I(\bX;\bS)$ into two terms \cite{shwartz2024compress}:
\begin{align}
    I(\bX;\bS) = \underbrace{I(\bX;\bS|\bY)}_\text{redundant information} + \underbrace{I(\bS;\bY).}_\text{task-related information}
\label{eq:informational-bottleneck}
\end{align}
This decomposition aligns with the information bottleneck principle \cite{tishby2000information, 10436784}, where the goal is to learn a representation that retains task-relevant information $I(\bS;\bY)$, while discarding irrelevant information $I(\bX;\bS|\bY)$. The contrastive learning approaches achieve this by maximizing the information in the representation relevant to the downstream task $I(\bS;\bY)$. 

\subsection{Redundant Information from the Causal Structure Model}\label{sec:causal-structure-model}
As stated in Eq.~\eqref{eq:informational-bottleneck}, the goal in reducing redundant information is to represent the compressed data specifically in terms of task-related information $I(\bS;\bY)$. To relax this relationship, we utilize the structured causal model (SCM) \cite{2022-IL-DIR}. 
\begin{definition}[Structured Causal Model]
Given an input image $\bX$, ground truth label $\bL$, causal part $\bC=I(\bS;\bL)$ (task-related representation), and non-causal part $\bV=I(\bX;\bS|\bL)$ (non-task-related representation), SCM can be represented as follows:
\begin{itemize}
    \item $\bC \rightarrow \bX \leftarrow \bV$: The input image $\bX$ consists of two disjoint parts: The causal part $\bC$ and the non-causal part $\bV$.
    \item $\bC \rightarrow \bL$: The causal part $\bC$ is the only endogenous parent to determining the ground-truth label $\bL$.
    \item $\bC \nleftrightarrow \bS$: This indicates the probabilistic dependencies between $\bC$ and $\bV$. There are spurious correlations between $\bC$ and $\bV$, therefore $\bC \nvDash \bV$.
\end{itemize}
\end{definition}
By separating data into causal and non-causal components, we hypothesize that the necessity to diminish redundant information can be alleviated by prioritizing the learning of representations pertaining to the causal components.  As indicated in \cite{2021-IL-SSDA}, the causal representations can be simplified by obtaining an invariant representation. To this end, we consider the following definition,
\begin{definition}[Symmetries]
    Let $\mathcal{G}$ be a group acting measurably on $\bS$. The conditional distribution $P(\bS\vert \bX)$ of $\bS$ given $\bX$ when $\mathcal{G}-$invariant if $\bS\vert \bX \overset{d}{=}\bS\vert \mathcal{T}\cdot \bX$:
    \begin{align}
        P(\bS\vert \bX)=P(\bS\vert \mathcal{T}\cdot \bX), \quad \forall \mathcal{T}\in \mathcal{G}
    \end{align}
    where $\bX$ and $\bS$ represent the source image and invariant representations, respectively.
\label{prop:symmetries}
\end{definition}
Following this definition, an invariant representation remains unchanged regardless of the permutation of input data. Consequently, we employ multi-view learning as an effective approach to generate data permutations. This enables us to discover an invariant representation that adheres to the condition of symmetry.

{Contrastive learning, despite its effectiveness in many contexts, faces notable challenges in device-to-device communication scenarios, where resource constraints and real-time requirements are paramount. A key limitation is its high computational cost, particularly in terms of memory consumption and training time \cite{wu2021rethinking}, which can overwhelm the limited processing power available on such devices. Furthermore, the method relies heavily on a large number of negative samples, images that do not contain the object of interest, to function effectively. However, generating these negative samples is often impractical in resource-constrained and time-sensitive environments.}

To address the inherent limitations of contrastive learning, we introduce a covariance-based method for feature extraction \cite{bardes2021vicreg}. This approach involves computing the covariance of the semantic features $\bS$ as follows:
\begin{align}\label{eq:covariance metho}
\mathbf{C}(\bS) = \frac{1}{K-1} \sum_{i=1}^{K}(\bS_i - \bar{\bS})(\bS_i - \bar{\bS})^T
\end{align}
where $\bar{\bS} = \frac{1}{K}\sum_{i=1}^{K}\bS_i$. This approach offers several advantages, particularly in the context of IoT communications. It inherently demands fewer computational resources, owing to its streamlined mathematical operations that minimize the need for extensive negative sample generation. Furthermore, by emphasizing variance and covariance among features, it naturally mitigates feature collapse, fostering a more uniform distribution of similar information across the embedding space. This characteristic plays a pivotal role in generating distinct and discriminative representations, thereby enhancing performance in downstream tasks. 
Section \ref{section:proposed_method} explains how the SC-GIR framework implements this covariance-based principle through a cross-correlation loss function applied to standardized latent representations.

\begin{figure*}[!htbp]
    \centering
    \includegraphics[width=.9\textwidth]{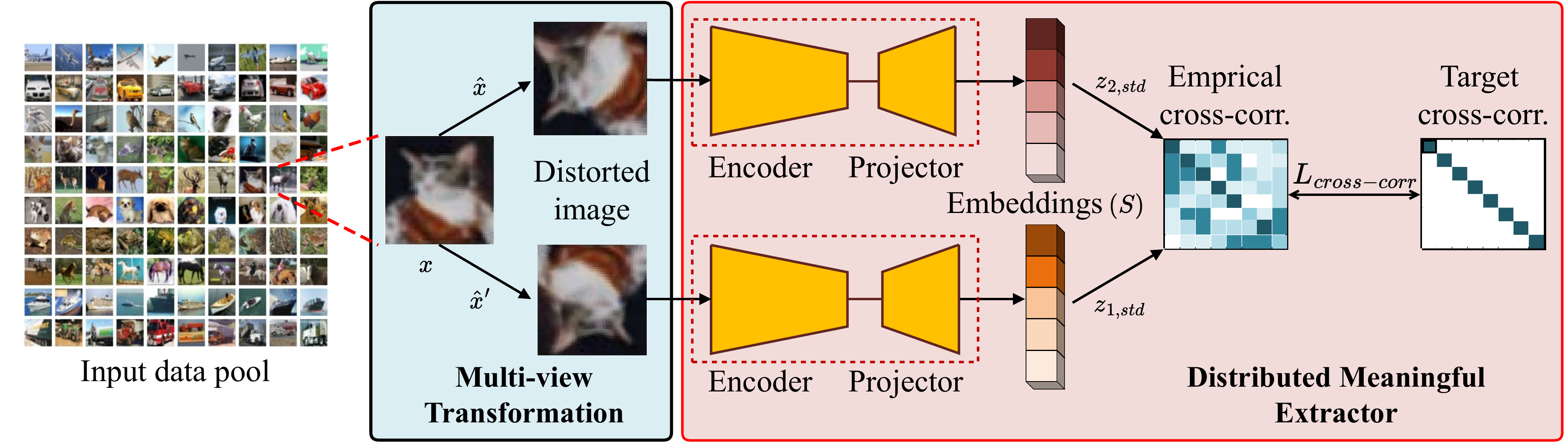}
    \caption{The proposed semantic encoder architecture consisting of the Multi-view transformation to generate the distorted view and distributed meaningful extractor to extract semantics.}
    \label{fig:encoder architecture}
\end{figure*}

\subsection{Channel Encoder and Decoder} 
In Fig.~\ref{fig:system_model}, the latent representation obtained from the semantic encoder is processed by the channel encoder. The channel encoder introduces redundancy into the data to bolster its resilience against channel perturbations. This redundancy-enriched channel encoded latent representation, denoted as ${\hat{\boldsymbol{s}}}$,  is then transmitted through the channel. At the receiver end, the data received as ${\boldsymbol{y}}$, representing the signal after experiencing channel perturbations, is fed into the channel decoder. The decoder's role is pivotal: it reconstructs the transmitted latent representation from ${\boldsymbol{y}}$, effectively mitigating the effects of channel perturbations and ensuring the integrity of the transmitted information. By employing algorithms tailored to error correction and information retrieval, the decoder plays a crucial role in maintaining communication reliability and quality.

\subsection{The Objective Design}
The proposed downstream task leverages the latent representation generated by the semantic encoder to perform classification, serving as a means to evaluate the effectiveness of our method. In this setup, the noise-corrupted latent vector, denoted by ${\boldsymbol{y}}$ and obtained after channel transmission, is fed into a classifier consisting of $m$ linear layers.

The objective of the goal-oriented AI (GOAI) task is to attain high classification accuracy despite the noise induced by the channel in the latent representation. This is accomplished by minimizing the GOAI loss, denoted as $\mathcal{L}_{\GOAI}$, which is the categorical cross-entropy loss quantifying the disparity between the predicted distribution of GOAI and the ground-truth labels $l$:
\begin{align}
     \mathcal{L}_{\mathrm{\GOAI}} =  -\frac{1}{N}\sum_{n=1}^{N}\sum_{c=1}^{C} l_{\textrm{nc}}\log(\hat{l}_{\textrm{nc}} ;w)
\end{align}
where $C$ denotes the number of classes, $N$ is the total number of samples, and $l_{nc}$ represents the ground truth in a one-hot encoded format for each sample $n$ and class $c$. The prediction $\hat{l}_{nc} = f_{\GOAI}({\boldsymbol{y}}; w)$ indicates the model's output for sample $n$ being in class $c$, where $f_{\GOAI}$ embodies the classifier's function, parameterized by the weight $w$.

The weight $w$ of the linear layers is updated through stochastic gradient descent (SGD) as
\begin{align}
    {w^{t+1}} \leftarrow {w^t} - \eta \cdot \nabla_{w} \mathcal{L}_{\GOAI}
\end{align}
where $\eta$ is the learning rate and $\nabla_{w} \mathcal{L}_{\GOAI}$ is the loss with respect to the weight.

To showcase the versatility of the proposed approach across various datasets, we adapt the configuration of the final layer according to the specific requirements of different datasets. Section~\ref{sec:Experiments} provides comprehensive details on the experiments conducted, demonstrating the performance and applicability of the proposed semantic communication framework in accomplishing GOAI tasks.

\section{Proposed SC-GIR Framework}\label{sec_Proposed}
\label{section:proposed_method}
To support the meaningful extractor described in Section~\ref{sec:causal-structure-model}, a semantic feature extractor is required as a central component for learning redundancy-reduced latent representations. To this end, we propose SC-GIR, a training-stage framework designed to learn both a semantic encoder and decoder.
In particular, SC-GIR consists of two key modules: the Multi-view Transformation and the Distributed Meaningful Extractor, as shown in Fig.~\ref{fig:encoder architecture}. Overall, our architecture leverages the Multi-view Transformation to generate diverse views of the data, enabling the encoder to learn shared, invariant features across these perturbations through the use of the cross-correlation loss $\mathcal{L}_{\mathrm{cross-corr}}$. This approach distinguishes our method from existing data transformation techniques, where transformations are typically trained jointly without specifically targeting invariant representation learning.

\subsection{Multi-view Transformation}
\begin{figure}[t]
    \centering
    \includegraphics[width=0.45\textwidth]{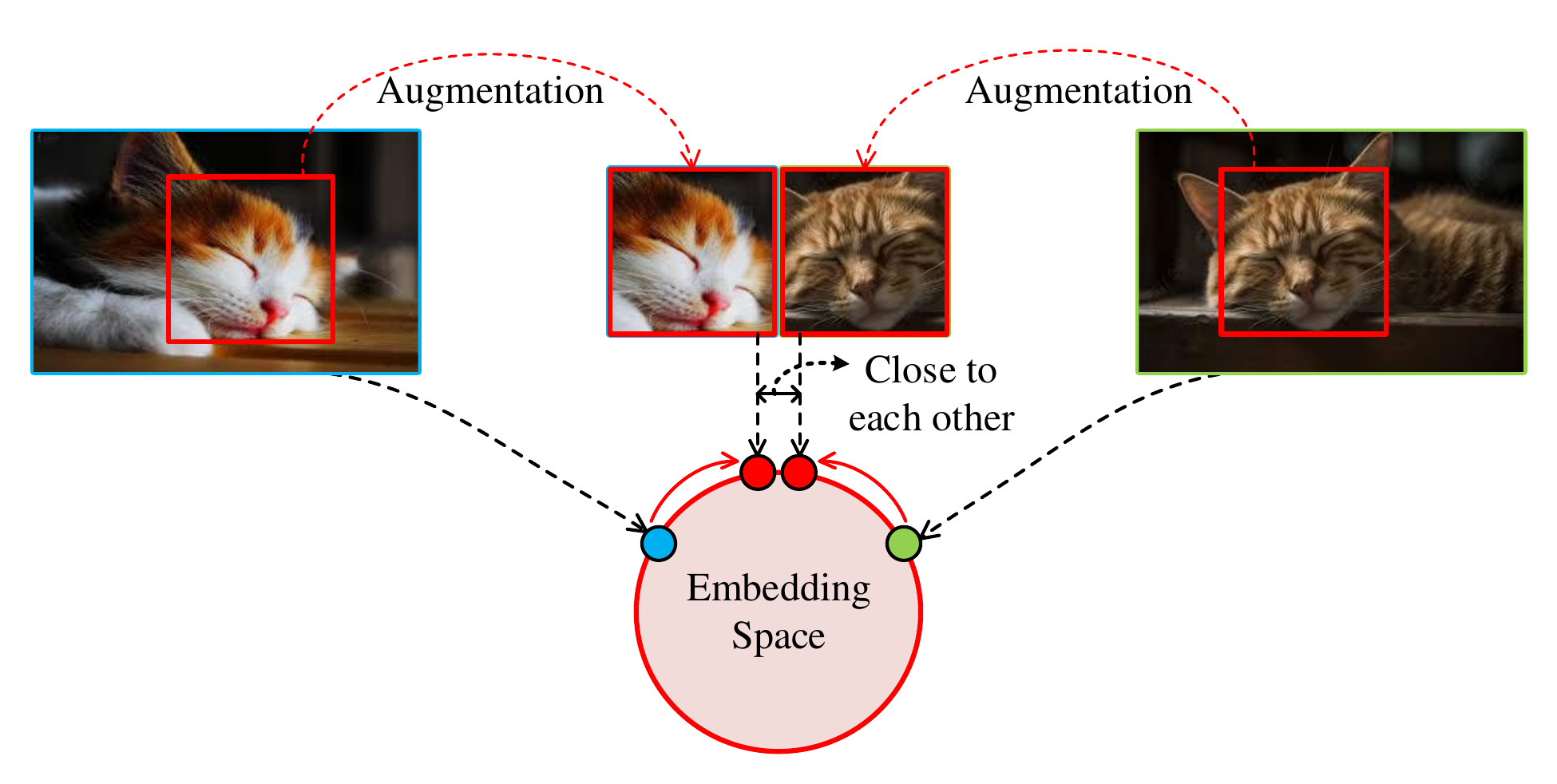}
    \caption{The learning process of the semantic encoder.}
    \label{fig:learning_process}
\end{figure}

\begin{figure}[t]
    \centering
    \includegraphics[width=0.45\textwidth]{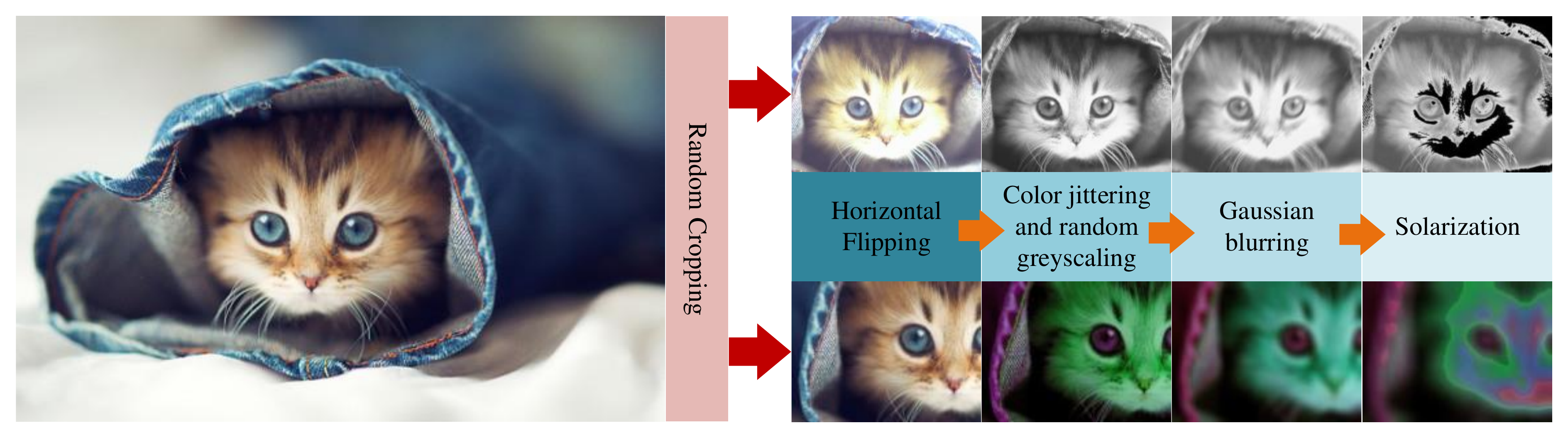}
    \caption{The data augmentation pipeline.}
    \label{fig:augmentation_scheme}
\end{figure}

Building on \cite{zbontar2021barlow, 2022-DG-CIRL, 2024-FL-FCCL}, the proposed multi-view transformation module illustrated in Fig.~\ref{fig:learning_process} acts as a stochastic data augmentation technique, generating two correlated contrastive views of a given data sample. Our approach emphasizes learning representations by contrastively comparing positively and negatively related pairs. This approach captures latent representations by treating each as a class vector within a higher-dimensional parametric space. We sequentially combine distortion transformation methods for image processing, resulting in two contrastive views of the input data. To integrate the multi-view architecture within the SC-GIR framework, our system preprocesses the input data $\boldsymbol{x}$ to generate two augmented views using the following augmentation regime: \textit{Random cropping}, \textit{horizontal flipping}, \textit{color processing}, \textit{Gaussian noise blurring}, \textit{solarization} as in Table~\ref{tab:Mul_SC-GIR} \cite{chen2020simple}. The overall multi-view transformation, along with the sequence of augmentation function application, is illustrated in Fig.~\ref{fig:augmentation_scheme}.


The process of generating two views is defined as
\begin{align}\label{eq:augmentation}
    &\boldsymbol{x}\xrightarrow[]{\mathcal{F_T}}\hat{\boldsymbol{x}},\quad
    \boldsymbol{x}\xrightarrow[]{\mathcal{F_T}^{\prime}}\hat{\boldsymbol{x}}^{\prime}
\end{align}
 where $\hat{\mathbf{x}}$ and $\hat{\mathbf{x}}^{\prime}$ are the outputs from a multi-view transformation module. Herein, $\mathcal{F_T}$ and $\mathcal{F_T}^{\prime}$ represent the cumulative stochastic augmentation functions with values specified in  Table~\ref{tab:Mul_SC-GIR}.

\noindent\textbf{Random cropping}: This entails selecting a random patch from the image, where the area is uniformly sampled with a crop probability $p_{\mathrm{crop}}$  relative to the original image's size. The chosen patch is subsequently resized to the target dimensions of $m \times m$ using bicubic interpolation.

\noindent\textbf{Horizontal flipping}: This operation allows for the optional flipping of the image horizontally with probability $p_{\mathrm{flip}}$. \vspace{5pt}

\noindent\textbf{Color processing}: This contains the following two steps.
\begin{itemize}
    \item Color jittering: In this process, adjustments are made to the brightness, contrast, saturation, and hue of the image by applying a uniformly random offset to all pixels. The application of these adjustments is chosen randomly for each selected patch.
    \item Color dropping: This operation optionally converts the image to grayscale. The intensity for each pixel $(\mathbf{r}, \mathbf{g}, \mathbf{b})$ after each application is determined by its luma component, calculated as $p_r\mathbf{r}+ p_g\mathbf{g} + p_b\mathbf{b}$.   
\end{itemize}\vspace{5pt}

\noindent\textbf{Gaussian noise blurring}: A square Gaussian kernel of size $m_{\mathrm{G}} \times m_{\mathrm{G}}$ is applied to the transformed image. The standard deviation of the kernel is randomly chosen from a uniform distribution over the interval $[0.1,\ 1]$.\vspace{5pt}

\noindent\textbf{Solarization}: This involves an optional color transformation for pixels with values in the range $[0,\ 1]$, which is defined as:
\begin{align}
    x \mapsto x \cdot \mathbbm{1}_{\{x<0.5\}} + (1-x) \cdot \mathbbm{1}_{\{x \geq 0.5\}}.
\end{align}

The distortion transformations 
$\mathcal{F_T}$ and $\mathcal{F_T}^{\prime}$ in Eq.~\eqref{eq:augmentation}, are compositions of the aforementioned image augmentations applied in the listed order, each with a predetermined probability. The parameters for these image distortion transformations are detailed in Table~\ref{tab:Mul_SC-GIR}.

It is important to emphasize that this Multi-view Transformation module is employed exclusively during the offline training phase. The goal is to enable the encoder to learn robust and invariant features by exposing it to diverse data views. During the online inference stage for real-time communication, a single, non-augmented input image is passed through the trained encoder only once to generate the latent representation for transmission. This single-pass approach ensures minimal computational overhead and latency, making the SC-GIR framework suitable for practical deployment on resource-constrained M2M devices.

\subsection{Distributed Meaningful Extractor}\label{sec:dme}
The main objective of the semantic encoder is to extract invariant representations from the shared views generated by the parallel augmentation function, thereby capturing the most meaningful mutual information within those views. Our aim is to enhance the capturing of mutual information by identifying invariant features within the data, which have been previously decomposed from the multi-view architecture, and aggregating them in the latent space \cite{zbontar2021barlow}.  {To this end, we propose the SC-GIR loss function, defined as:
\begin{align}\label{eq:loss-function}
        \mathcal{L}_{\mathrm{cross-corr}} = \sum_{i}^{}\left ( 1 - \mathcal{C}_{ii} \right )^2 + \lambda\sum_{i}^{}\sum_{i \neq j}^{}\left ( \mathcal{C}_{ij}\right )^2
\end{align}where $\mathcal{C}$ denotes the empirical cross-correlation matrix between $\boldsymbol{z}_1$ and $\boldsymbol{z}_2$ which are the standardized outputs of two parallel encoders, $f_1$ and $f_2$, each consisting of multi-layer projection heads with similar architectures}. In particular, we have:
\begin{align}\label{eq:standard_output}
        &\boldsymbol{z}_{1,std} = f_{\mathrm{Stand}}\big(f_1(\hat{\boldsymbol{x}})\big)\\  
        &\boldsymbol{z}_{2,std} =  f_{\mathrm{Stand}}\big(f_2(\hat{\boldsymbol{x}}^\prime)\big).
\end{align}
The projection heads receive the encoder outputs and return $\boldsymbol{z}_i$, a $d$-dimensional vector. The $ f_{\mathrm{Stand}}$ function standardizes this latent vector utilizing the mean $\mu$ and the standard deviation $\sigma$ of that vector. Therefore, prior to calculating the empirical cross-correlation, we obtain $\boldsymbol{z}_{1,std} \in \mathbb{R}^{n\times d}$ and $\boldsymbol{z}_{2,std} \in \mathbb{R}^{n\times d}$ where $n$ is the number of samples.
Following that, the empirical cross-correlation matrix $\mathcal{C} = \frac{\boldsymbol{z}_{1,std}^T \boldsymbol{z}_{2,std}}{n} \in \mathbb{R}^{d\times d}$ is obtained using the standardised vectors. 
This cross-correlation loss function directly embodies the covariance-based redundancy reduction principle outlined in Eq. \eqref{eq:covariance metho}. By operating on standardized outputs, it effectively computes the normalized covariance Eq. \eqref{eq:loss-function}, providing a stable and interpretable objective for promoting both feature invariance and de-correlation.
The parameter $\lambda$  in Eq.~\eqref{eq:loss-function} represents the scaling factor between the diagonal and off-diagonal losses for the total loss.

The first component of the cross-correlation focuses on the off-diagonal loss, aiming to maximize shared information between the two outputs by establishing an invariant representation through the preceding transformation regime. The subsequent off-diagonal term is dedicated to redundancy reduction.
 
Intuitively, the diagonal elements of the cross-correlation matrix should approach $+1$, reflecting the maximization of mutual information, while the off-diagonal elements should approach $0$, signifying the reduction of redundancy. These constraints ensure that the latent representation undergoes semantic compression, optimizing its suitability for subsequent downstream tasks. Our objective function aims to encourage the clustering of different data samples in the same latent space while simultaneously pulling apart dissimilar latent spaces, as depicted in Fig.~\ref{fig:learning_process}.
The learning process of the semantic encoder can be explained from the perspective of the information bottleneck principle \cite{wang2022rethinking, tishby2015deep}. The latent representation of data ${\bX}$ obtained via the encoder should be sufficiently informative for the downstream task ${\bY}$. Therefore, the model aims to concatenate a greater amount of non-shared invariant information within the latent representation ${\bS}$, which is consistent across the two views, thereby increasing the mutual information $I(\bS;\bY)$.

However, the information bottleneck principle also underscores the significance of a succinct representation. Hence, the mutual information between the original data $\bX$, and the latent representation $\bS$, denoted by ${I}(\bX;\bS)$, should be minimized to ensure a compressed and efficient representation. Following the information bottleneck principle ($\mathcal{IB}$), the latent representation should strike a balance between these two competing objectives. To formalize this trade-off, the objective function $\mathcal{IB}$ can be mathematically formulated as
\begin{align}\label{eq:information_bottleneck}
        \max_{p(t|x)}\ \mathcal{IB} = {I}\left (\bY;\bS  \right ) - \alpha {I}\left (\bX;\bS\right )
\end{align}
where $\alpha$ is the scaling factor. Given an input $\bX$ and a desired output $\bY$, the model learns a representation $T$ that captures the information essential for fulfilling task $\bY$. Through simplifications and approximations\cite{zbontar2021barlow}, the objective function is reformulated into our loss function in Eq.~\eqref{eq:loss-function}, which underlies the minimally sufficient representation obtained via the semantic encoder.
\begin{algorithm}[t]
\fontsize{9}{9}\selectfont
    \caption{Semantic Encoder Algorithm for the Proposed SC-GIR Framework}
    \label{alg:encoder_algorithm}
    \SetKwFunction{isOddNumber}{isOddNumber}
    % \SetKwInput{Input}{Input}
    % \SetKwInput{Output}{Output}
    \SetKwInOut{KwIn}{Input}
    \SetKwInOut{KwOut}{Output}
    \KwIn{Data samples aligned serially which are to be semantically encoded;\\ $\mathcal{B}$ is batch size with $b$ samples.}
    \textbf{Initialize:} The neural network parameters: $\boldsymbol{\theta}$ for $f_{\boldsymbol{\theta}}$, $f_{\mathrm{diag}}$ which vectorizes the diagonal elements of a square matrix.
    
    %\KwOut{Processed list.}
    \For{batch $b$ sampled from $N$ samples}{
        $y_1, y_2 = a(x)$;\tcp*[f]{generate two views}

        $\boldsymbol{z}_1 = f_{\boldsymbol{\theta}}(y_1);\ \boldsymbol{z}_2 = f_{\boldsymbol{\theta}}(y_2)$;\tcp*[f]{obtain the latent representation}
        \BlankLine
        $\boldsymbol{z}_{\mathrm{1,norm}}= \frac{\boldsymbol{z}_1 - \mathbb{E}[\boldsymbol{z}_1]}{\sigma_{\boldsymbol{z}_1}};\ \boldsymbol{z}_{\mathrm{2,norm}}= \frac{\boldsymbol{z}_2 - \mathbb{E}[\boldsymbol{z}_2]}{\sigma_{\boldsymbol{z}_2}}$;\tcp*[f]{standardisation}
        
        $\mathcal{C} = \frac{\sum_{b} \boldsymbol{z}_{\mathrm{1,norm}} \cdot \boldsymbol{z}_{\mathrm{2,norm}}}{\sqrt{\sum _b\left ( \boldsymbol{z}_{\mathrm{1,norm}}^2 \right )}\sqrt{\sum_{b}\boldsymbol{z}_{\mathrm{2,norm}}^2}}$;\tcp*[f]{cross-correlation matrix}
        \BlankLine
        $\mathcal{L}_{\mathrm{on}} = \sum_{\mathcal{C}_{ii}}\big(f_{\mathrm{diag}}(\mathcal{C}) - 1\big)^2$;\tcp*[f]{summation over the diagonal elements}
    
        $\mathcal{L}_{\mathrm{off}} = \sum_{\mathcal{C}_{ij}} \big(\mathcal{C} - f_{\mathrm{diag}}(\mathcal{C}) + 1\big)^2$;\tcp*[f]{summation over the off-diagonal elements}
       
        $\mathcal{L}_{\mathrm{cross-corr}} = \mathcal{L}_{\mathrm{on}} + \mathcal{L}_{\mathrm{off}}$;\\
    }
\KwOut{$\mathcal{L}_{\mathrm{cross-corr}},\ \boldsymbol{z}_1$.}
\end{algorithm}

The entire training process is succinctly captured in Algorithm ~\ref{alg:encoder_algorithm}, where a batch $\mathcal{B}$ of data is processed through the semantic encoder to yield two distinct views. These views are then used to compute the cross-correlation matrix, as outlined in line 6. Subsequently, both diagonal and weighted off-diagonal losses are determined, leading to the computation of the total loss and the extraction of the latent representation. The primary objective throughout the training phase is to minimize the training loss, denoted as $\mathcal{L}_{\text{cross-corr}}$.

\textit{Discussion on model complexity:}
The proposed SC-GIR method operates during the training phase and does not alter the model architecture used for inference. Consequently, it incurs no additional inference latency, memory footprint, or computational cost (\textit{e.g.} FLOPs) compared to existing approaches such as SemCC and DeepJSCC. This design ensures that SC-GIR remains suitable for deployment in resource-constrained environments, such as IoT and wireless edge devices, while offering improved training effectiveness through enhanced representation learning.

\section{Experimental Evaluations}\label{sec_Numerical}
In this section, we first provide an overview of the datasets used for the experiments. Subsequently, we delve into the implementation details of each component within the considered system model. Following this, we present our experimental results, categorized into two main evaluations: the encoder assessment and the analysis related to the GOAI task. These results are accompanied by a thorough comparative analysis aimed at assessing the efficacy of our proposed methodology.

\subsection{Experimental Setup}
\label{sec:Experiments}

{\bf Datasets}: In accordance with standard practices, we assess the performance of SC-GIR using established datasets, including CIFAR-10, CIFAR-100 \cite{krizhevsky2009learning}, EMNIST \cite{cohen2017emnist}, FMNIST \cite{xiao2017fashion}, STL-10 \cite{coates2011analysis}, Flower-17 \cite{1640927}, Cityscape \cite{cordts2016cityscapes} and PACS \cite{li2017deeper}. These datasets are widely acknowledged for their quality and consistency in evaluating a variety of computer vision classification algorithms.

\begin{table}[t]
    \centering
    \caption{Augmentation Functions and Corresponding Parameters of the SC-GIR Multi-view Transformation Module for Each View as Described in Eq.~\eqref{eq:augmentation}}
    \label{tab:Mul_SC-GIR}
    \resizebox{0.95\columnwidth}{!}{
    \begin{tabular}{l|cc}
        \hline \text {\textbf{Functions} } & $\mathcal{F_T}$ & $\mathcal{F_T}^{\prime}$ \\
        \hline\hline \text {Random crop probability } & 1.0 & 1.0 \\
        \text{Horizontal flip probability } & 0.5 & 0.5 \\
        \text{Brightness adjustment max intensity } & 0.4 & 0.4 \\
        \text{Contrast adjustment max intensity } & 0.4 & 0.4 \\
        \text{Saturation adjustment max intensity } & 0.2 & 0.2 \\
        \text{Hue adjustment max intensity } & 0.1 & 0.1 \\
        \text{Greyscaling probability} & 0.2 & 0.2 \\
        \text{Gaussian blurring probability } & 1.0 & 0.1 \\
        \text{Solarization probability } & 0.0 & 0.2 \\
        \hline
    \end{tabular}
    }
\end{table}

For example, the CIFAR dataset consists of 60,000 images (50,000 for training and 10,000 for testing), distributed across either 10 or 100 different classes, respectively. To prepare these datasets, we employ a multi-view augmentation pipeline that generates two distorted views for each data point. This pipeline involves resizing the images to a standardized size of $32 \times 32$ pixels and applying a series of random transformations with specified probabilities. The transformations include horizontal flipping, color jittering, grayscale conversion, Gaussian blurring, and solarization, as outlined in Table~\ref{tab:Mul_SC-GIR}.

\begin{table}[t]
\caption{Neural Network Architecture of the Semantic Encoder}
\label{tab:Neural_architecture}
\centering
\begin{tabular}{l|l|l}

\hline
\textbf{Network}                                        & \textbf{Layer Name} & \textbf{Properties}  \\ \hline
\multirow{9}{*}                       & conv1    & $3 \times 3$, 64, stride 1                                                                       \\
                                               & pool1      & $3 \times 3$, max pool, stride 2                    \\
                                               &     \vspace{0.01pt}       &  \vspace{0.01pt}       \\
                                               & conv2\_x    & $\begin{bmatrix}
                                                                3 \times 3, & 64 \\
                                                                3 \times 3, & 64 
                                                                \end{bmatrix} \times 3$    \\
                                                &     \vspace{0.01pt}       &  \vspace{0.01pt}       \\
                                                
            \textbf{Encoder}                                    & conv3\_x    & $\begin{bmatrix}
                                                                3 \times 3, & 128 \\
                                                                3 \times 3, & 128 
                                                                \end{bmatrix} \times 4$    \\
                                                &     \vspace{0.01pt}       &  \vspace{0.01pt}       \\
                                                & conv4\_x    & $\begin{bmatrix}
                                                                3 \times 3, & 256 \\
                                                                3 \times 3, & 256 
                                                                \end{bmatrix} \times 6$    \\
                                                &     \vspace{0.01pt}       &  \vspace{0.01pt}       \\
                                                & conv5\_x    & $\begin{bmatrix}
                                                                3 \times 3, & 512 \\
                                                                3 \times 3, & 512 
                                                                \end{bmatrix} \times 3$    \\  \hline
\multirow{3}{*}{\textbf{MLP}} & {layer 1}    & {[}512, 512{]}, GELU, Batchnorm1d  \\
                     
                                                & {layer 2}    & {[}512, 1024{]}, GELU, Batchnorm1d \\
            
                                                 & {layer 3}    & {[}1024, 2048{]}, GELU, Batchnorm1d \\ \hline

\end{tabular}
\end{table}

{\bf Encoder}: We utilize ResNet-34 \cite{he2016deep} as the backbone architecture for all the following experiments. To accommodate the integration of multi-layer projection heads, we adapt the backbone architecture by replacing the final classification layer of ResNet with a multi-layer projection network (MLP), as depicted in Table~\ref{tab:Neural_architecture}. The projection head consists of three linear layers with output dimensions 512, 1024, and 2048, respectively. Following each linear layer, we apply a Gaussian Error Linear Unit (GELU) activation function \cite{hendrycks2016gaussian} due to its smoothness and differentiability across all input ranges, which is essential for effective training. Additionally, batch normalization is applied after each GELU activation to further stabilize and accelerate the learning process.
During training, we set the trade-off parameter $\lambda$ to $5\times10^{-4}$. The encoder is optimized using the Adam optimizer \cite{kingma2014adam} with a mini-batch size of 64 and a learning rate of $10^{-4}$.

\begin{table}[t]
\centering
\caption{Neural Network Architecture of GOAI Task}
\label{tab:GOAI_architecture}
\begin{tabular}{l|l|l}
\hline
\textbf{Network}               & \textbf{Layer Name} & \textbf{Properties}                                                                               \\ \hline
\multirow{4}{*} & layer 1    & \begin{tabular}[c]{@{}l@{}}{[}512, 1024{]}, \\ PReLU, BatchNorm1d\end{tabular}        \\
                    &     \vspace{0.01pt}       &  \vspace{0.01pt}       \\
                      & layer 2    & \begin{tabular}[c]{@{}l@{}}{[}1024, 1024{]}, \\ PReLU, BatchNorm1d\end{tabular}       \\
           \textbf{GOAI}           &     \vspace{0.01pt}       &  \vspace{0.01pt}       \\
                      & layer 3    & \begin{tabular}[c]{@{}l@{}}{[}1024, 1024{]}, \\ PReLU, BatchNorm1d\end{tabular}       \\
                      &     \vspace{0.01pt}       &  \vspace{0.01pt}       \\
                      & layer 4    & \begin{tabular}[c]{@{}l@{}}{[}512, output\_dim{]}, \\ PReLU,  BatchNorm1d\end{tabular} \\ \hline
\end{tabular}
\end{table}

{\bf Goal-oriented AI Task (GOAI)}: To assess the versatility of the proposed semantic encoder across a range of downstream tasks, we employ classification as our primary evaluation framework. The GOAI architecture consists of four linear layers, each activated by a Parametric Rectified Linear Unit (PReLU) \cite{he2015delving}. The output dimensions of these layers vary depending on the dataset used, as outlined in Table~\ref{tab:GOAI_architecture}. To improve training stability and accelerate learning, 1D batch normalization layers are inserted after each linear layer. Additionally, we employ a learning rate scheduler that utilizes a cosine annealing strategy \cite{loshchilov2016sgdr} in conjunction with the Adam optimization algorithm \cite{kingma2014adam} to facilitate efficient convergence.

{
{\bf Baseline schemes}: To evaluate the effectiveness of the proposed SC-GIR framework, we compare its performance against several established traditional and semantic communication approaches. 
\begin{itemize}
\item \textit{Deep Joint Source-Channel Coding (DeepJSCC)} \cite{bourtsoulatze2019deep}: This end-to-end semantic communication model employs an encoder-decoder architecture for simultaneous source and channel coding. The output of DeepJSCC is subsequently passed through a pre-trained ResNet-18 \cite{7780459} for the goal-oriented classification task.

\item \textit{Contrastive Learning-based Semantic Communication (SemCC)} \cite{10530261}: SemCC leverages contrastive learning techniques to improve the alignment between transmitted and received semantics.

\item \textit{Semantic Re-Encoding (SemRE)} \cite{10530261}: This approach introduces an auxiliary semantic encoder to guide training when the downstream task or application model is unavailable during transmission.

\item \textit{Transformer-based Semantic Communication (DeepSC)} \cite{9398576}: A semantic communication framework built upon the transformer architecture, designed to preserve task-relevant features across noisy channels.

\item \textit{Traditional Communication Systems (BPG + LDPC)}: Represented by conventional source and channel coding techniques, such as Bit-Perceptual Grouping (BPG) for source coding combined with LDPC codes for channel protection.
\end{itemize}
To ensure a fair comparison, all methods are evaluated on the same downstream task using their respective reconstructed outputs. This evaluation setup allows us to specifically assess the impact of different semantic encoder architectures on overall system performance.
}

{\bf Metrics}: We consider various performance metrics, gauging both feature quality and downstream task impact. Cosine similarity assesses the encoder's ability to produce informative, less redundant, and stable representations. High cosine similarity between the original data and encoded representations indicates effective semantic capture with reduced redundancy. For task performance, we use the F1 score to compare our method against the baseline in the GOAI task, serving as the primary metric.

\subsection{Performance Comparison with Baselines}
\begin{table*}[!hbt]
\caption{Performance of SC-GIR Compared to Several Baselines Across Different Datasets}
\label{tab:baseline_performance}
\resizebox{\linewidth}{!}{%
\begin{tabular}{lcccccccc}

\toprule
\textbf{Algorithm} & \textbf{CIFAR-10} & \textbf{CIFAR-100} & \textbf{MNIST} & \textbf{STL-10} & \textbf{FMNIST} & \textbf{Flower-17} & \textbf{Average} \\
\midrule
\textbf{SC-GIR} & $\mathbf{87.2}\pm 0.1$ & $\mathbf{98.0} \pm 0.0$ & $\mathbf{99.3} \pm 0.4$ & $\mathbf{85.5} \pm 0.2$ & $86.5 \pm 0.3$ & $73.1 \pm 1.8$ & 66.6 \\
SemCC & $87.2 \pm 0.1$ & $ 74.7 \pm 0.1$ & $99.1 \pm 0.6$ & $73.5 \pm 0.8$ & $95.3 \pm 2.2$ & $75.6\pm 0.8$ & 65.4 \\
SemRE& $ 83.5 \pm 1.0$ & $ 65.0 \pm 0.0$ & $98.4 \pm 0.7$ & $69.6 \pm 0.2$ & $73.5 \pm 0.7$ & $65.2 \pm 1.2$ & 65.6 \\
DeepJSCC & $ 61.0 \pm 1.2$ & $64.0 \pm 0.0$ & $95.4 \pm 0.6$ & $71.9 \pm 0.8$ & $78.8 \pm 0.3$ & $70.9 \pm 0.8$ & 66.7 \\
DeepSC & $53.5 \pm 0.1$ & $56.0 \pm 0.0$ & $86.4 \pm 0.2$ & $58.9 \pm 1.0$ & $69.3 \pm 0.3$ & $60.7 \pm 0.9$ & 66.7 \\
BPG+2/3 rate LDPC+16QAM & $85.5 \pm 0.1$ & $94.1 \pm 0.2$ & $78.5 \pm 0.0$ & $54.5 \pm 0.3$ & $90.6 \pm 0.6$ & $80.2 \pm 1.6$ & 63.8 \\
BPG+1/2 rate LDPC+16QAM & $75.5 \pm 0.2$ & $85.9 \pm 0.0$ & $76.0 \pm 0.2$ & $47.3 \pm 0.4$ & $85.9 \pm 0.6$ & $75.7 \pm 1.5$ & 65.6 \\
BPG+3/4 rate LDPC+4QAM & $72.5 \pm 0.3$ & $78.8 \pm 0.1$ & $68.6 \pm 0.6$ & $59.8 \pm 0.4$ & $86.9 \pm 0.6$ & $70.7 \pm 1.5$ & 66.1 \\
BPG+2/3 rate LDPC+4QAM & $58.7 \pm 0.1$ & $80.8 \pm 0.1$ & $65.6 \pm 0.3$ & $52.5 \pm 0.7$ & $80.6 \pm 0.9$ & $65.8 \pm 1.3$ & 65.5 \\
BPG+1/2 rate LDPC+4QAM & $69.4 \pm 0.1$ & $66.0 \pm 0.0$ & $55.5 \pm 0.4$ & $49.3 \pm 0.5$ & $82.1 \pm 0.0$ & $70.7 \pm 1.1$ & 66.4 \\
\bottomrule
\end{tabular}%
}
\end{table*}

We conduct a comprehensive benchmarking analysis across six diverse datasets to evaluate the performance of the proposed SC-GIR approach. The model is compared against both semantic communication baselines (SemCC, SemRE, DeepJSCC, DeepSC) and traditional communication baselines (BPG paired with various LDPC and quadrature amplitude modulation (QAM) configurations), using classification accuracy as the primary performance metric. The results, summarized in Table~\ref{tab:baseline_performance}, highlight the effectiveness of SC-GIR. SC-GIR delivers outstanding performance on CIFAR-100, MNIST, and STL-10, consistently outperforming all baselines in both semantic and traditional settings. On CIFAR-10, SC-GIR remains highly competitive, matching the best-performing baseline. While its performance is comparatively lower on FMNIST and Flower-17, where it is surpassed by SemCC, DeepJSCC, and specific BPG configurations, it still maintains a strong overall showing. 

Notably, SC-GIR achieves an average accuracy of $66.6\%$ across all datasets, nearly matching the top-performing methods DeepJSCC and DeepSC (both at $66.7\%$). In addition to its solid average performance, SC-GIR demonstrates better consistency, exhibiting lower standard deviations on several datasets, particularly CIFAR-100, STL-10, and FMNIST, compared to other methods that show greater performance fluctuations. In summary, SC-GIR excels on complex datasets, remains competitive in more standard scenarios, and, despite minor dips on a few datasets, offers a compelling balance of accuracy and stability. This combination of reliable performance and robustness makes it a strong candidate for semantic communication applications, often providing a meaningful advantage over existing alternatives.

\subsection{Impact of Wireless Channel}
\label{wireless_channel_impact}

\begin{figure}[!t]
\centering
     \subfloat[Test accuracy versus the bandwidth compression ratio under AWGN channel.\label{fig:AWGN_test_acc}]{\includegraphics[width=\linewidth]{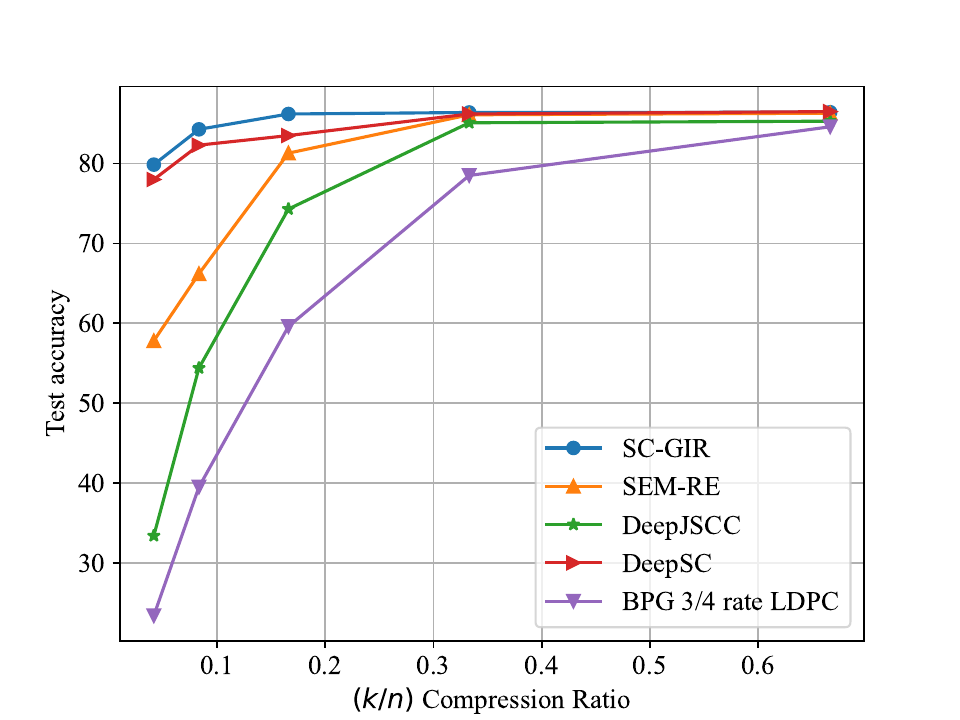}}\\
     \subfloat[Test accuracy versus the bandwidth compression ratio under the Rayleigh channel.\label{fig:Rayleigh_test_acc}]{\includegraphics[width=\linewidth]{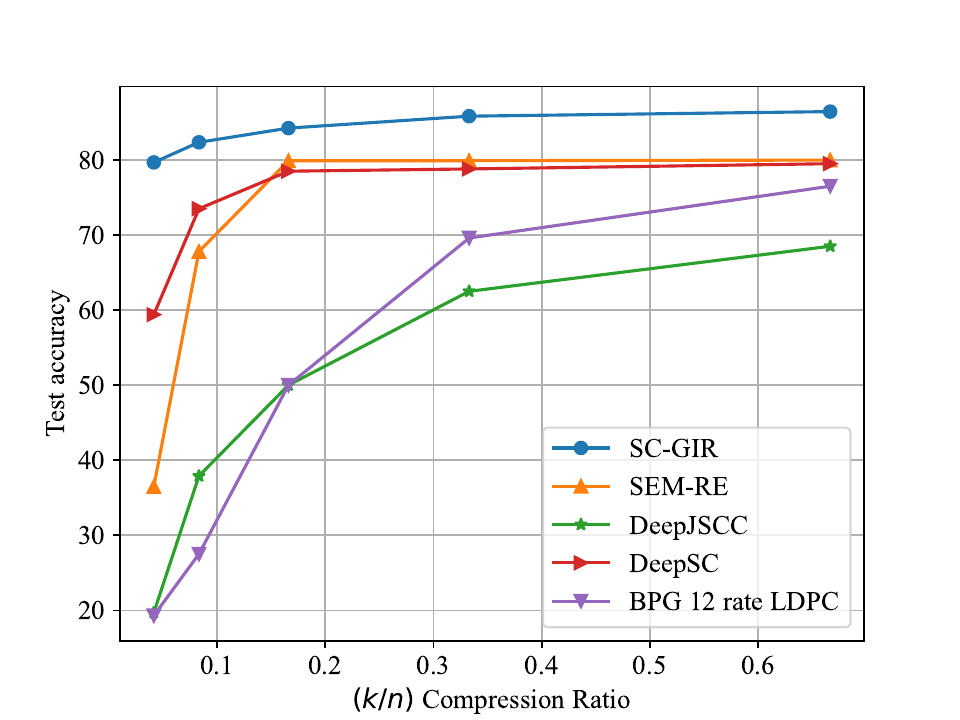}} 
    \caption{Impact of compression ratio on the CIFAR-10 dataset with SNR = $5$ dB.}
    \label{fig:snr_accuracy}
    %\vspace{-10pt}
\end{figure}

To evaluate the effectiveness of the proposed SC-GIR method under varying wireless channel conditions, we benchmark it against several established baselines, including SEM-RE~\cite{10530261}, DeepJSCC~\cite{2020-SemCom-DJSCCF}, DeepSC~\cite{xie2021deep}, and BPG with $3/4$ rate LDPC. Experiments are conducted on the CIFAR-10 dataset, where each image is 3,072 bytes. We evaluate a range of bandwidth compression ratios $(k/n)$ from 0.1 to 0.6, corresponding to transmitted data sizes between $307$ and $1,843$ bytes per image, tested under both AWGN and Rayleigh fading channels, each operating at a fixed signal-to-noise ratio (SNR) of $5$ dB. The resulting classification accuracy, shown in Fig.~\ref{fig:AWGN_test_acc} for AWGN and Fig.~\ref{fig:Rayleigh_test_acc} for Rayleigh, illustrates SC-GIR's performance under these conditions.

Under the AWGN channel, SC-GIR consistently outperforms all baseline methods. At a low compression ratio of $k/n = 0.1$, it achieves nearly $85\%$ accuracy, significantly outperforming SEM-RE ($70\%$), DeepJSCC ($65\%$), DeepSC ($80\%$), and BPG $3/4$ rate LDPC ($30\%$). As the compression ratio increases to $k/n = 0.6$, SC-GIR maintains its lead with an accuracy of $88\%$, compared to DeepSC ($85\%$), SEM-RE ($83\%$), DeepJSCC ($75\%$), and BPG 3/4 rate LDPC ($65\%$). These results emphasize SC-GIR's strong performance, especially under stringent bandwidth constraints. In Rayleigh fading channels, SC-GIR demonstrates notable robustness against channel impairments. At $k/n = 0.1$, it achieves $80\%$ accuracy, outperforming SEM-RE ($65\%$), DeepJSCC ($50\%$), DeepSC ($70\%$), and BPG 3/4 rate LDPC ($25\%$). At the higher ratio of $k/n = 0.6$, SC-GIR reaches $83\%$ accuracy, again surpassing DeepSC ($80\%$), SEM-RE ($78\%$), DeepJSCC ($65\%$), and BPG 3/4 rate LDPC ($50\%$). These findings highlight SC-GIR's resilience in more challenging and dynamic wireless environments.

Unlike competing methods that struggle under low-bandwidth or fading conditions, such as BPG $3/4$ rate LDPC, which drops to $25\%-30\%$ at $k/n = 0.1$, or DeepJSCC, which underperforms in Rayleigh channels, SC-GIR sustains high accuracy across both channel types and all compression levels. This consistent performance underscores its suitability for resource-constrained and variable wireless settings. Overall, SC-GIR's superior accuracy across scenarios at SNR $= 5$ dB, when compared with both traditional and learning-based baselines, confirms its strong potential for deployment in practical semantic communication systems.
\begin{figure}[!t]
\centering
     \subfloat[Test accuracy versus SNR with  $k/n=1/6$.\label{fig:test_acc_against_snr_16}]{\includegraphics[width=\linewidth]{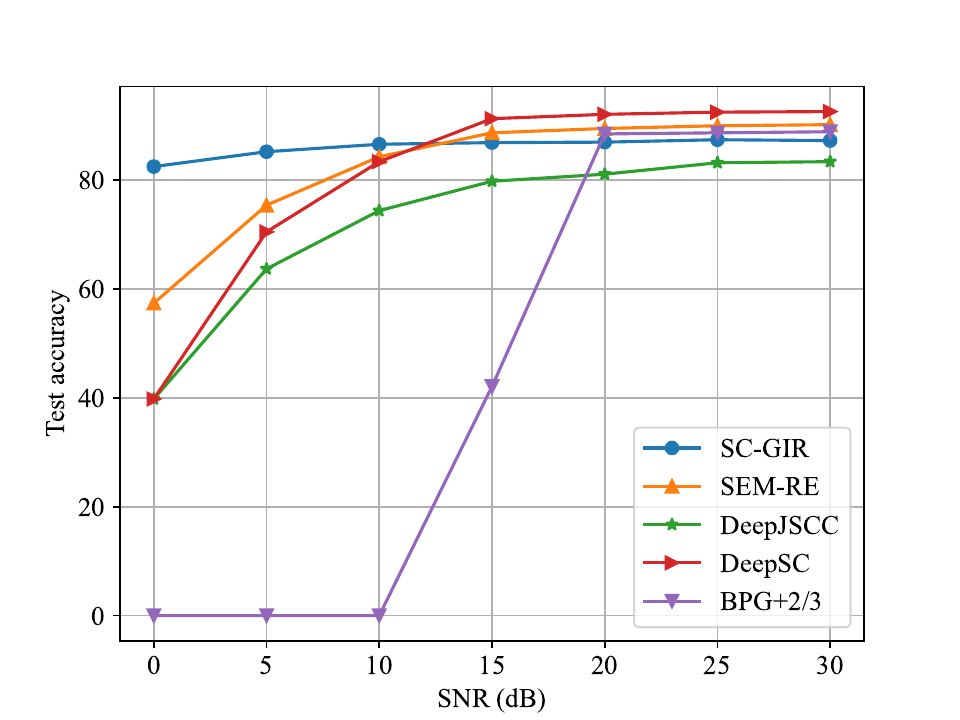}}\\
     \subfloat[Test accuracy  versus SNR with  $k/n=1/12$.\label{fig:test_acc_against_snr_12}]{\includegraphics[width=\linewidth]{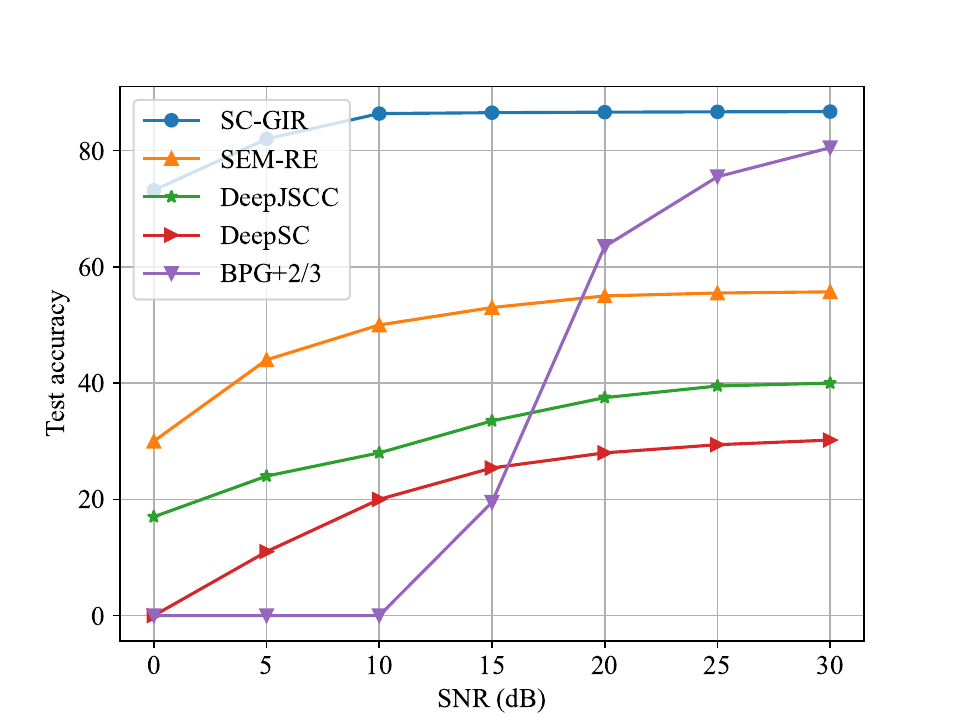}} 
    \caption{Impact of SNR under Rayleigh channel on the CIFAR-10 dataset for two compression ratios ($k/n=\{1/6,\,1/12\}$).}
    \label{fig:compression_accuracy}
\end{figure}

Fig.~\ref{fig:compression_accuracy} illustrates the performance of different SC methods over a Rayleigh fading channel, evaluated on the CIFAR-10 dataset across
various SNR levels and two compression ratios ($k/n=1/6$ and $k/n=1/12$). The results demonstrate the superior robustness and efficiency of SC-GIR under challenging wireless conditions. In Fig.~\ref{fig:test_acc_against_snr_16}, where the compression ratio is $k/n = 1/6$, SC-GIR achieves a test accuracy of $60\%$ at $0$ dB, significantly outperforming SEM-RE ($40\%$), DeepJSCC ($20\%$), and BPG+2/3 LDPC+16QAM (near $0\%$). As the SNR increases to $30$ dB, SC-GIR reaches $80\%$ accuracy, compared to $70\%$ for SEM-RE, $50\%$ for DeepJSCC, and $40\%$ for BPG+2/3. In the more constrained setting shown in Fig.~\ref{fig:test_acc_against_snr_12}, with a higher compression ratio of $k/n = 1/12$, SC-GIR maintains a test accuracy of $50\%$ at $0$ dB, while SEM-RE, DeepJSCC, and BPG+2/3 achieve only $30\%$, $15\%$, and $0\%$, respectively. At $30$ dB, SC-GIR still leads with $75\%$ accuracy, ahead of SEM-RE ($60\%$), DeepJSCC ($45\%$), and BPG+2/3 ($35\%$). Notably, SC-GIR experiences only a minimal performance degradation (approximately $5\%$) between the two compression settings at high SNR levels, highlighting its strong generalization and compression resilience. This robustness is attributed to SC-GIR’s hybrid design, which combines semantic communication principles with generative image restoration. By selectively preserving task-relevant features and suppressing noise-induced distortions, SC-GIR delivers consistent performance under both low-SNR and high-compression scenarios, making it a highly suitable solution for bandwidth-limited and noisy wireless communication environments.

\subsection{Encoder Evaluation}
\begin{figure}[t]
\centering
     \subfloat[CIFAR-10 \label{fig:TL_SCGIR_cifar}]{\includegraphics[width=0.5\linewidth]{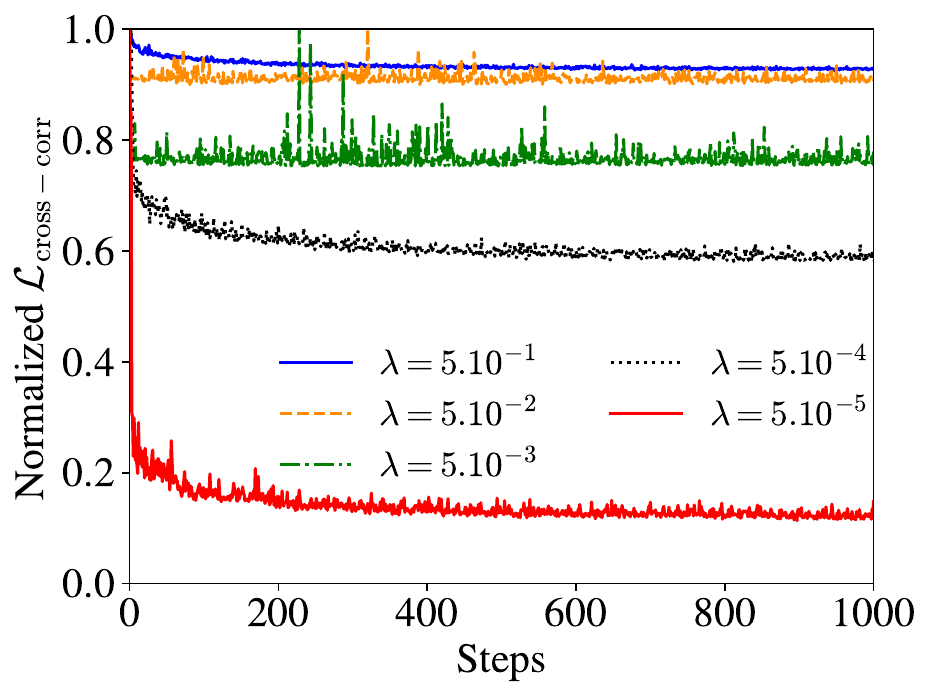}}
     \subfloat[FMNIST\label{fig:TL_SCGIR_FMNIST}]{\includegraphics[width=0.5\linewidth]{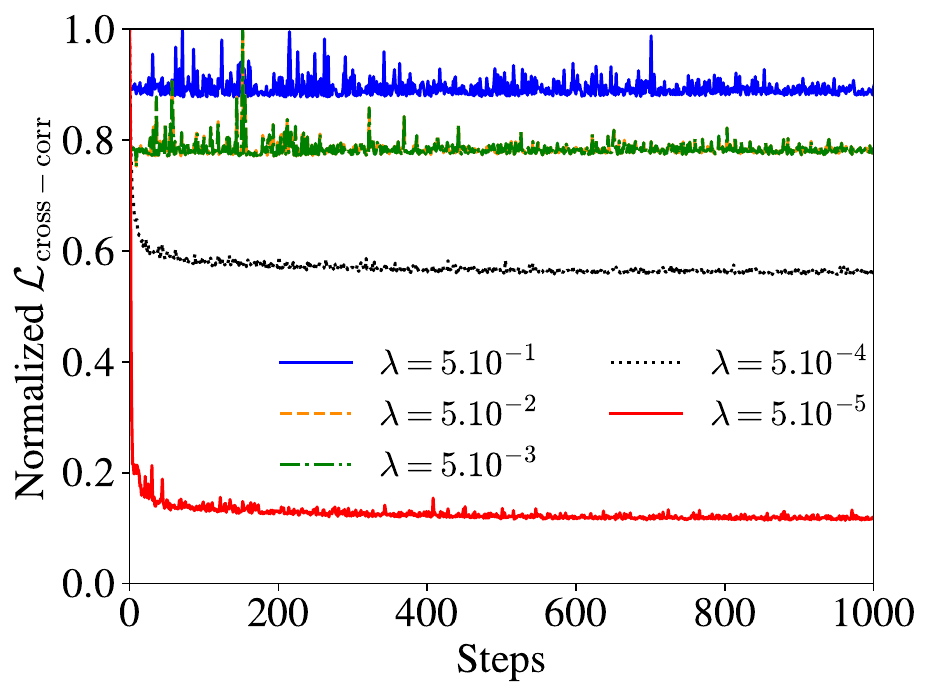}}\\
     \subfloat[MNIST\label{fig:TL_SCGIR_MNIST}]{\includegraphics[width=0.5\linewidth]{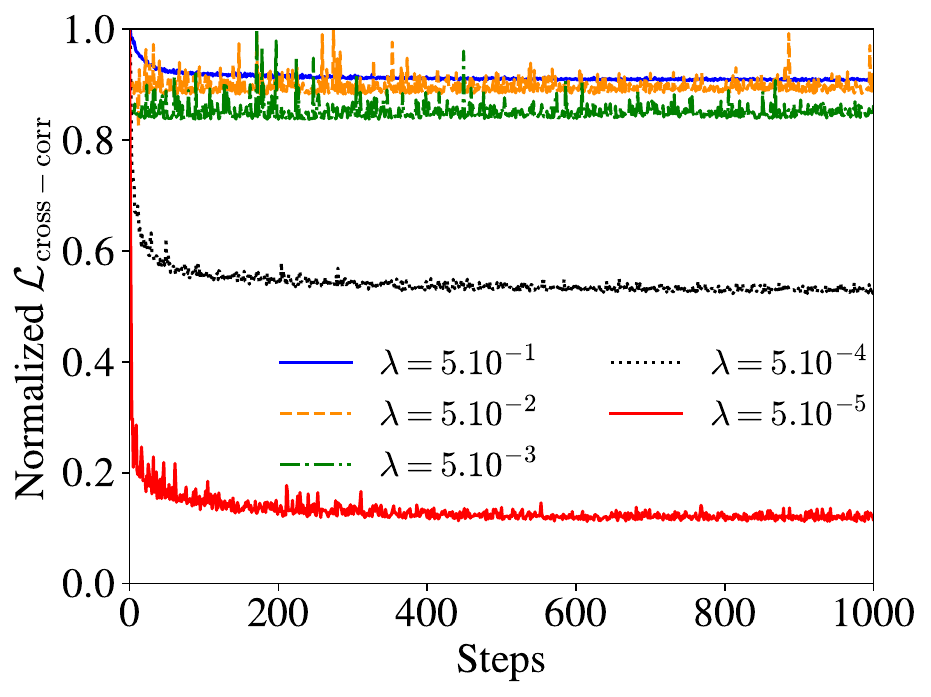}} 
     \subfloat[Flower-17 \label{fig:TL_SCGIR_Flower17}]{\includegraphics[width=0.50\linewidth]{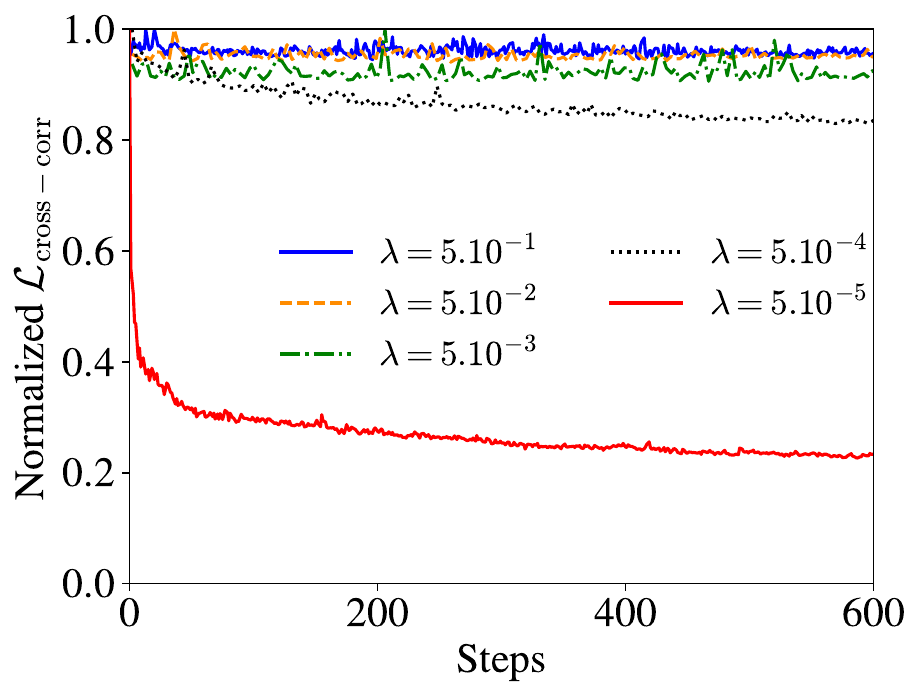}}\\
     \subfloat[CIFAR-100\label{fig:TL_SCGIR_CIFAR100}]{\includegraphics[width=0.5\linewidth]{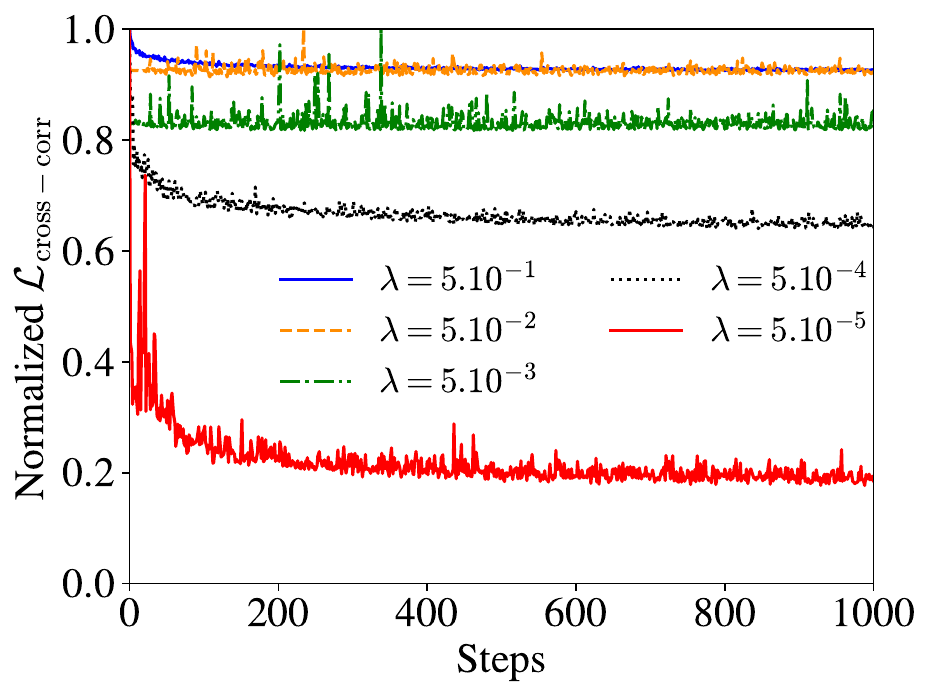}} 
     \subfloat[STL-10 \label{fig:TL_SCGIR_STl10}]{\includegraphics[width=0.5\linewidth]{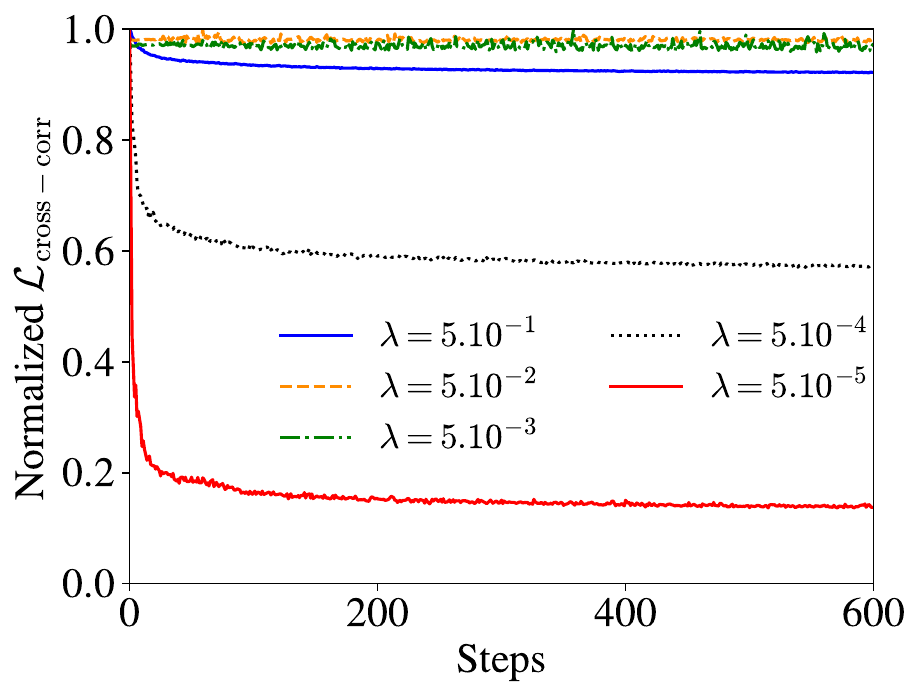}}
    \caption{The normalized SC-GIR loss function with different weighting coefficients $\lambda$ applied to the diagonal and off-diagonal components of the loss.}\label{fig:TL_SCGIR}
\end{figure}

In Fig.~\ref{fig:TL_SCGIR}, we analyze the effect of the scaling factor $\lambda$, which controls the trade-off between feature invariance and redundancy reduction. This analysis is crucial, as stable training convergence is essential for model robustness under noisy channel conditions. Across all six datasets, larger values of $\lambda$ (\textit{e.g.} $\lambda \geqslant  5\times10^{-4}$) result in unstable and non-converging training loss, indicating that an overly strong redundancy penalty disrupts learning. In contrast, setting $\lambda  = 5\times10^{-4}$ consistently yields fast, stable, and low-error convergence. This enables the model to learn coherent and meaningful representations, forming the basis for its robust performance across the various channel conditions evaluated in this study.

\begin{figure*}[t]
\centering
     \subfloat[CIFAR-10 \label{fig:CS_cifar}]{\includegraphics[width=0.29\linewidth]{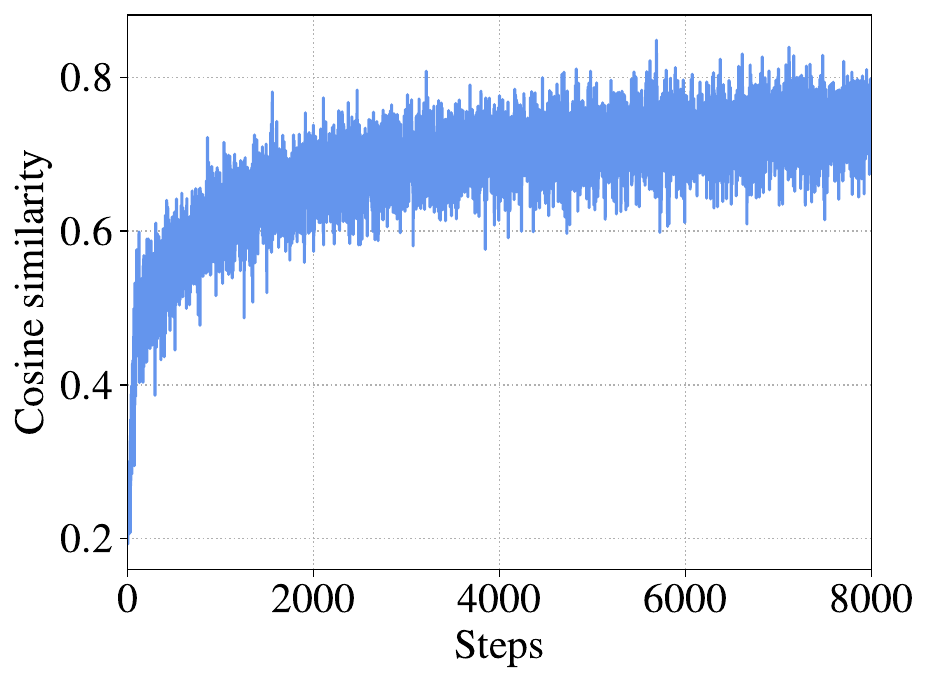}}
     \subfloat[FMNIST\label{fig:CS_FMINST}]{\includegraphics[width=0.28\linewidth]{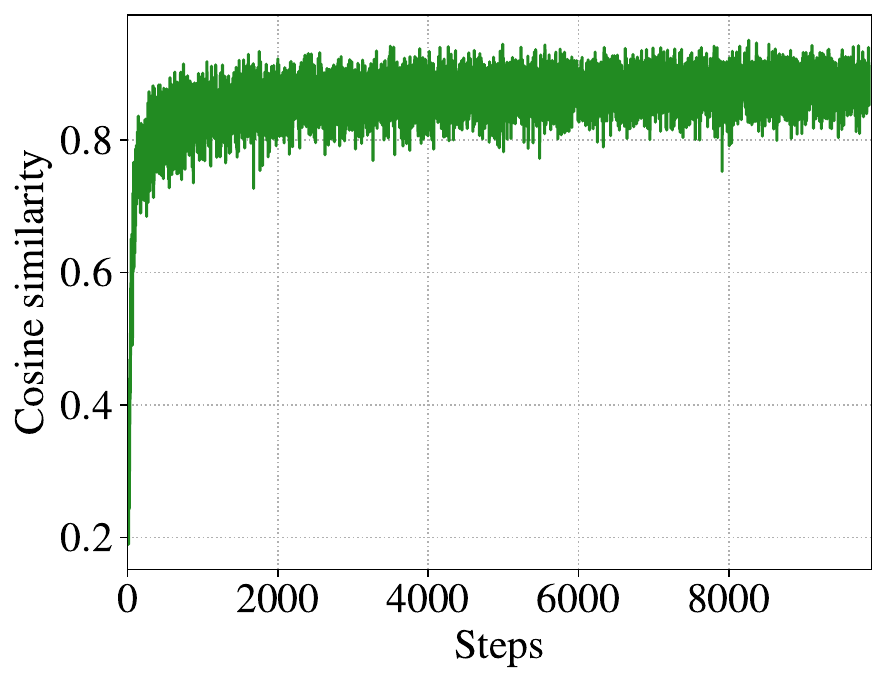}}
     \subfloat[EMNIST\label{fig:CS_MNIST}]{\includegraphics[width=0.28\linewidth]{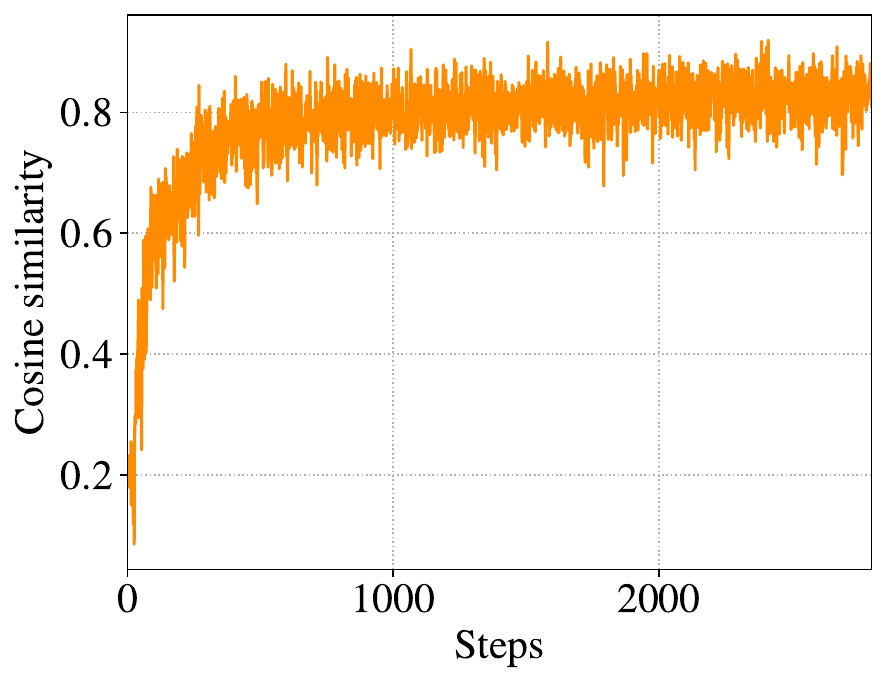}} \\
     \subfloat[Flower-17 \label{fig:CS_Flower17}]{\includegraphics[width=0.28\linewidth]{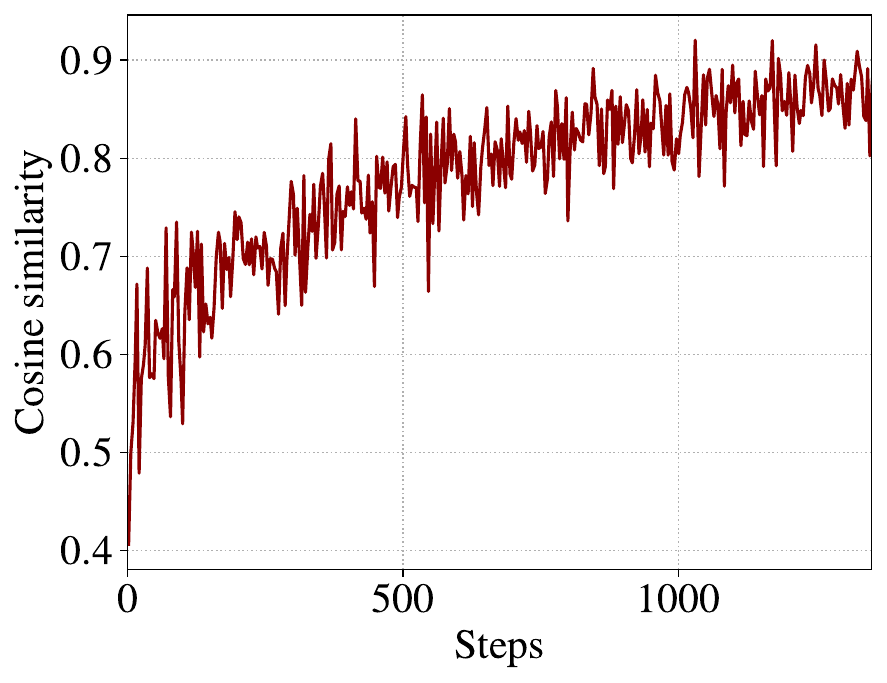}}
     \subfloat[CIFAR-100 \label{fig:CS_cifar100}]{\includegraphics[width=0.28\linewidth]{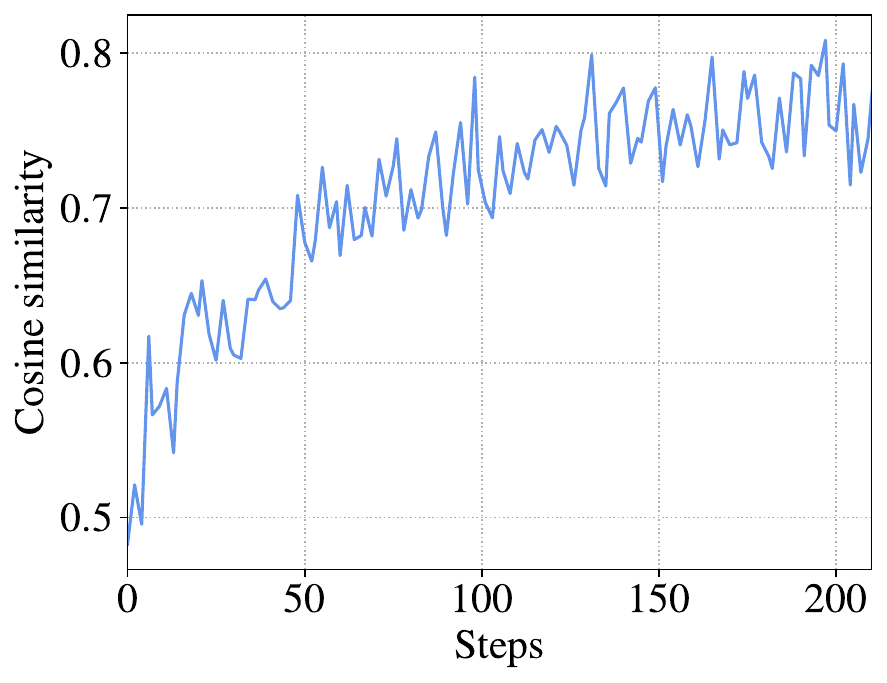}}
     \subfloat[STL-10\label{fig:CS_STL10}]{\includegraphics[width=0.28\linewidth]{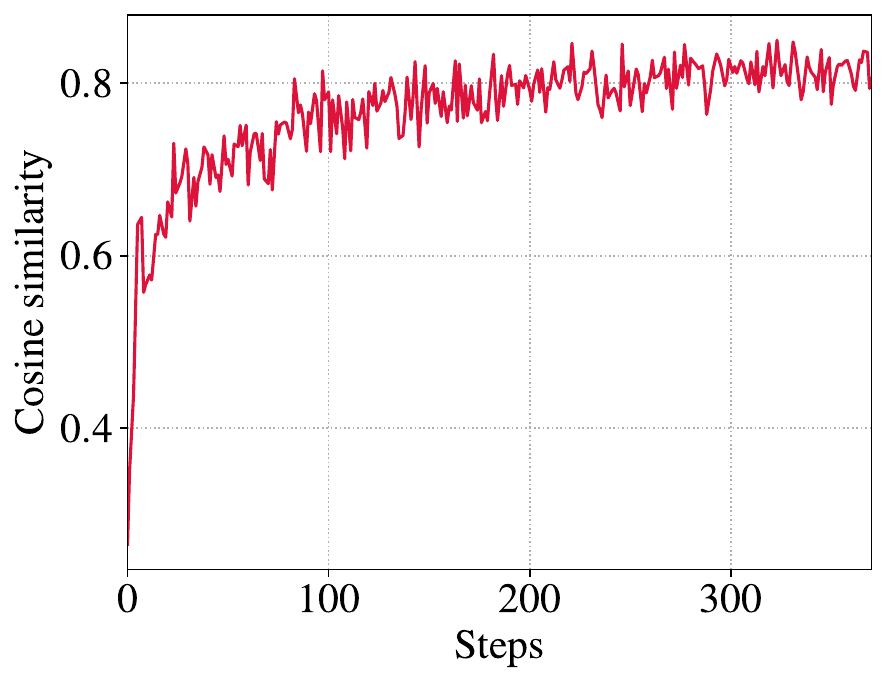}}
     
    \caption{Cosine similarity between the latent representations obtained via the semantic encoder during training.}\label{fig:CS}
\end{figure*}

Fig.~\ref{fig:CS} depicts the cosine similarity between latent representations derived from the proposed semantic encoder during training for different datasets. The noticeable increase in cosine similarity indicates the model's increasing resilience to augmentations applied to the two views. As a result, the encoded representations become progressively less susceptible to variations introduced through data augmentation, thereby capturing underlying semantic information more effectively.
The sharper incline in cosine similarity observed for FMNIST and MNIST compared to CIFAR-10 and Flower-17 suggests that the lower-dimensional nature of the data facilitates a swifter initial assimilation of the underlying semantic features.

\begin{figure*}[t]
\centering
     \subfloat[CIFAR-10 \label{fig:Senu04_1}]{\includegraphics[width=0.3\linewidth]{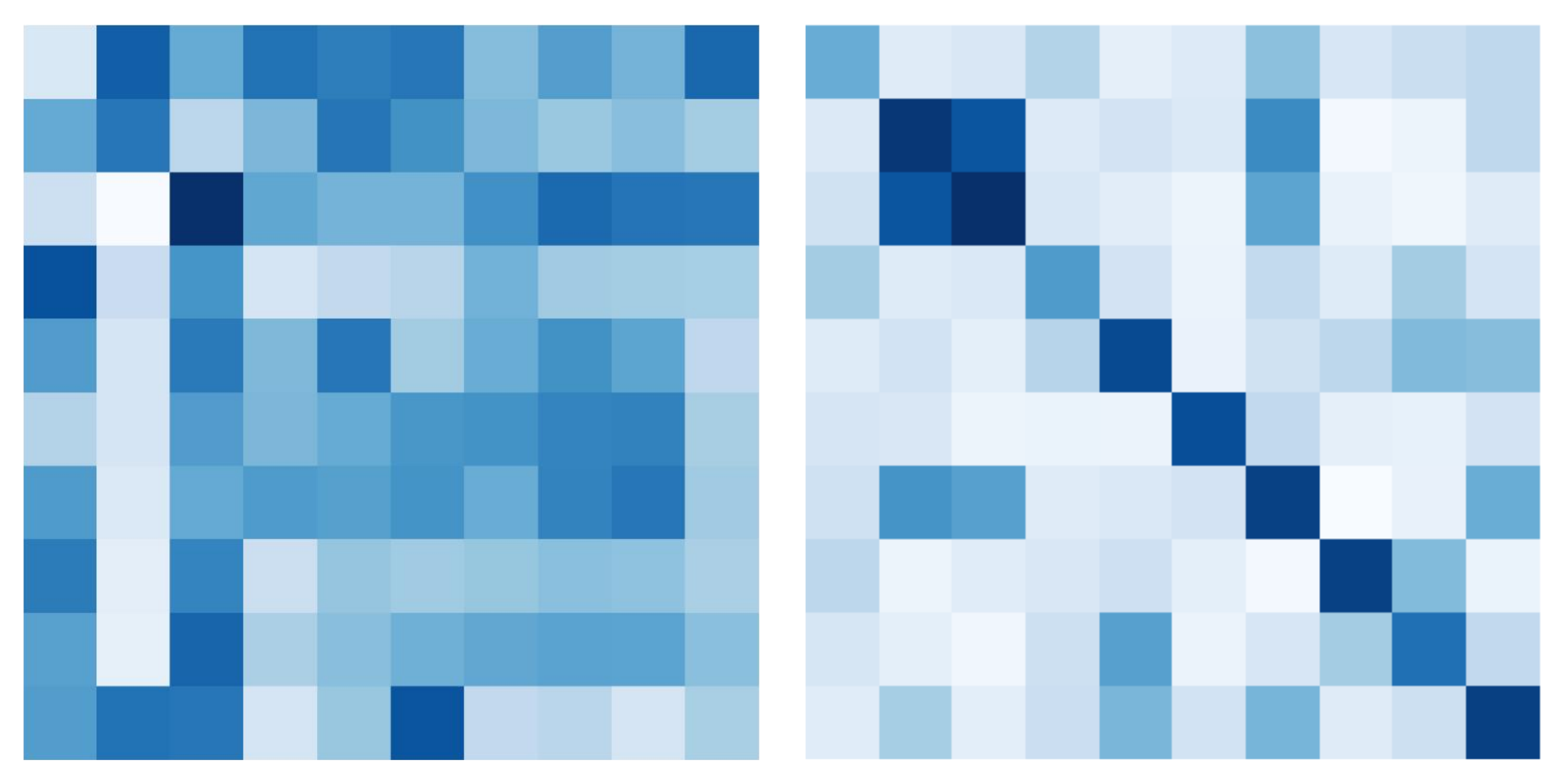}}
     \subfloat[FMNIST\label{fig:Senu04_3}]{\includegraphics[width=0.3\linewidth]{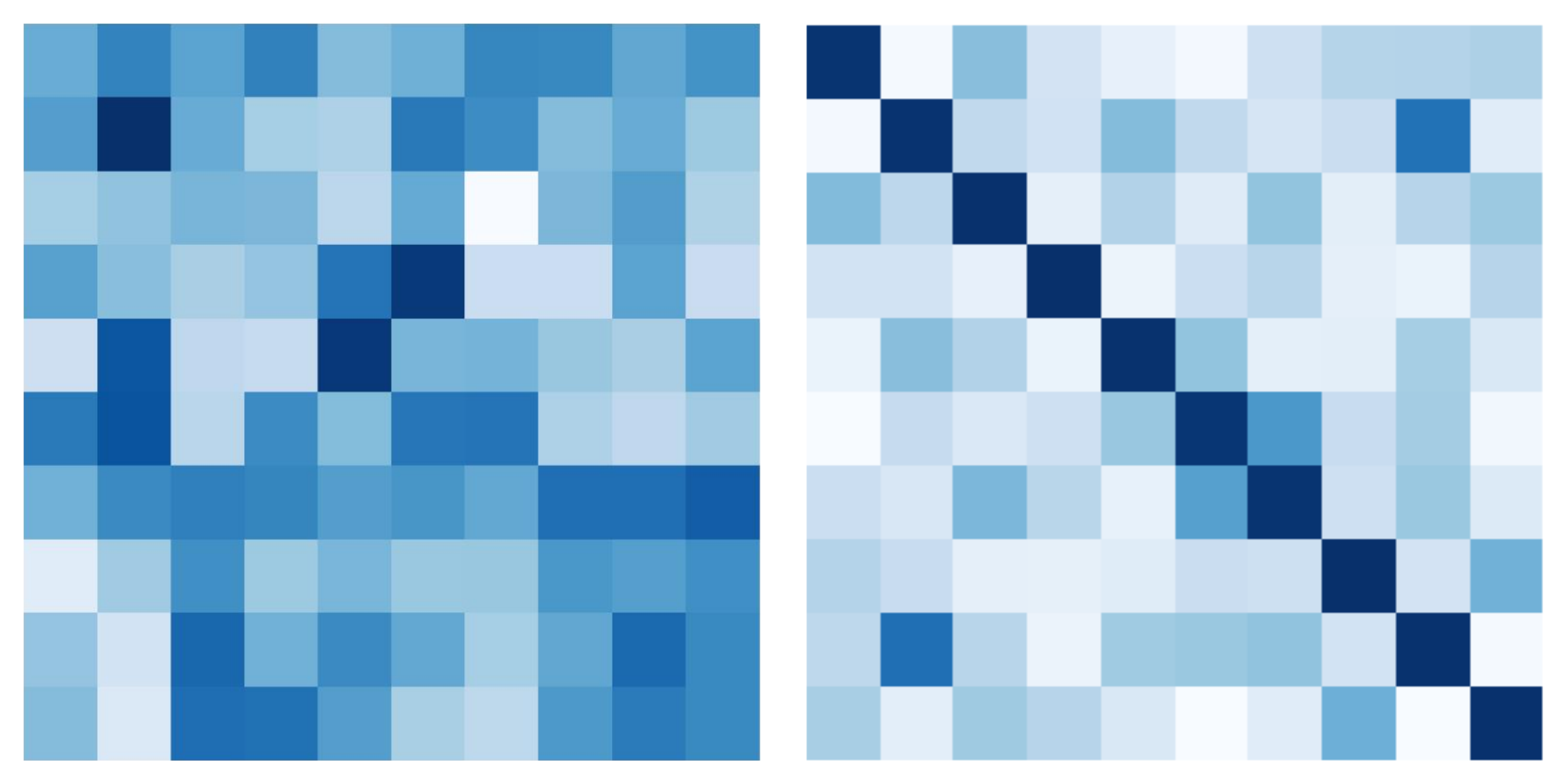}}
     \subfloat[MNIST\label{fig:Senu04_4}]{\includegraphics[width=0.3\linewidth]{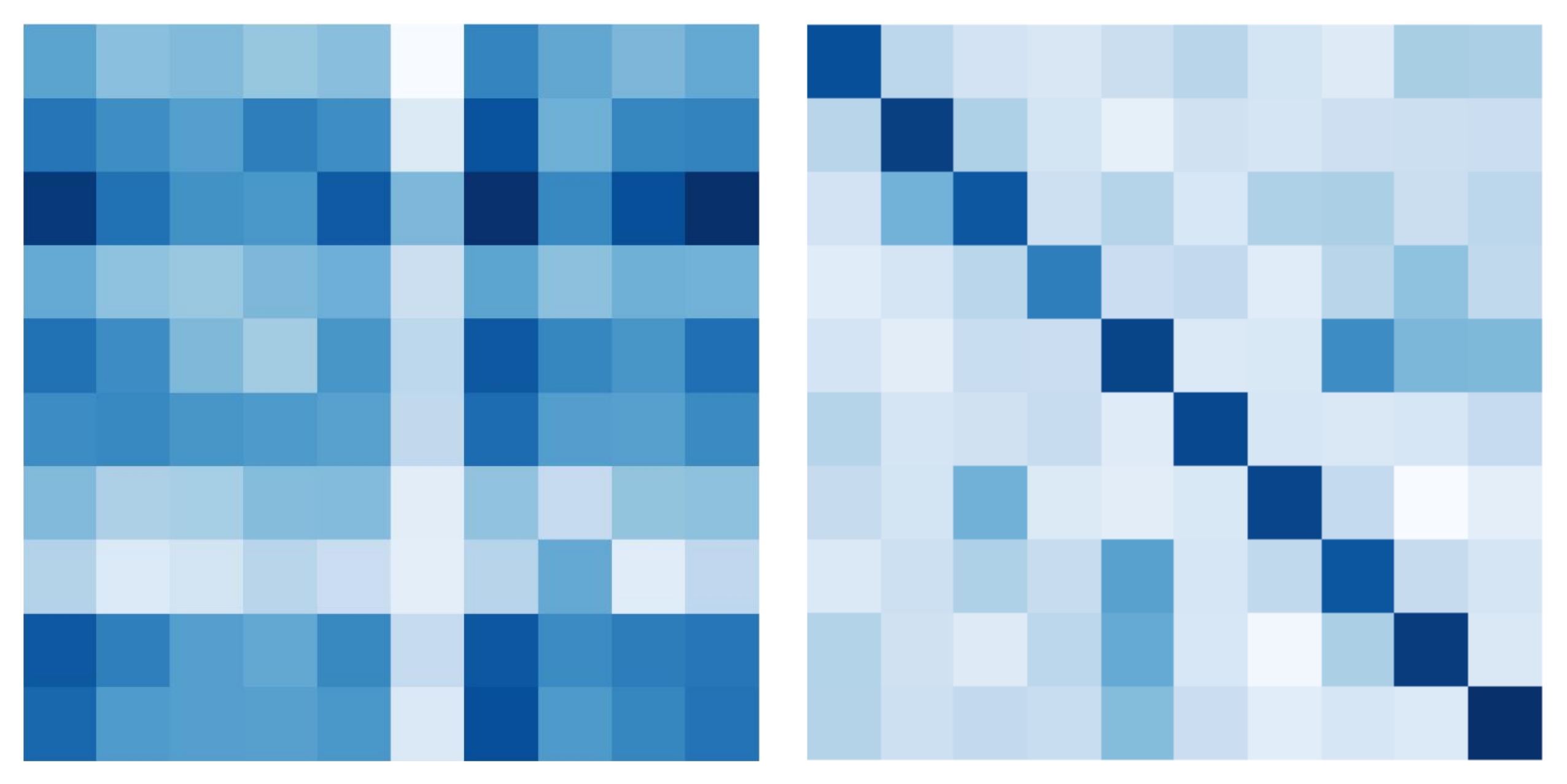}} \\
     \subfloat[CIFAR-100 \label{fig:cifar100}] 
     {\includegraphics[width=0.3\linewidth]{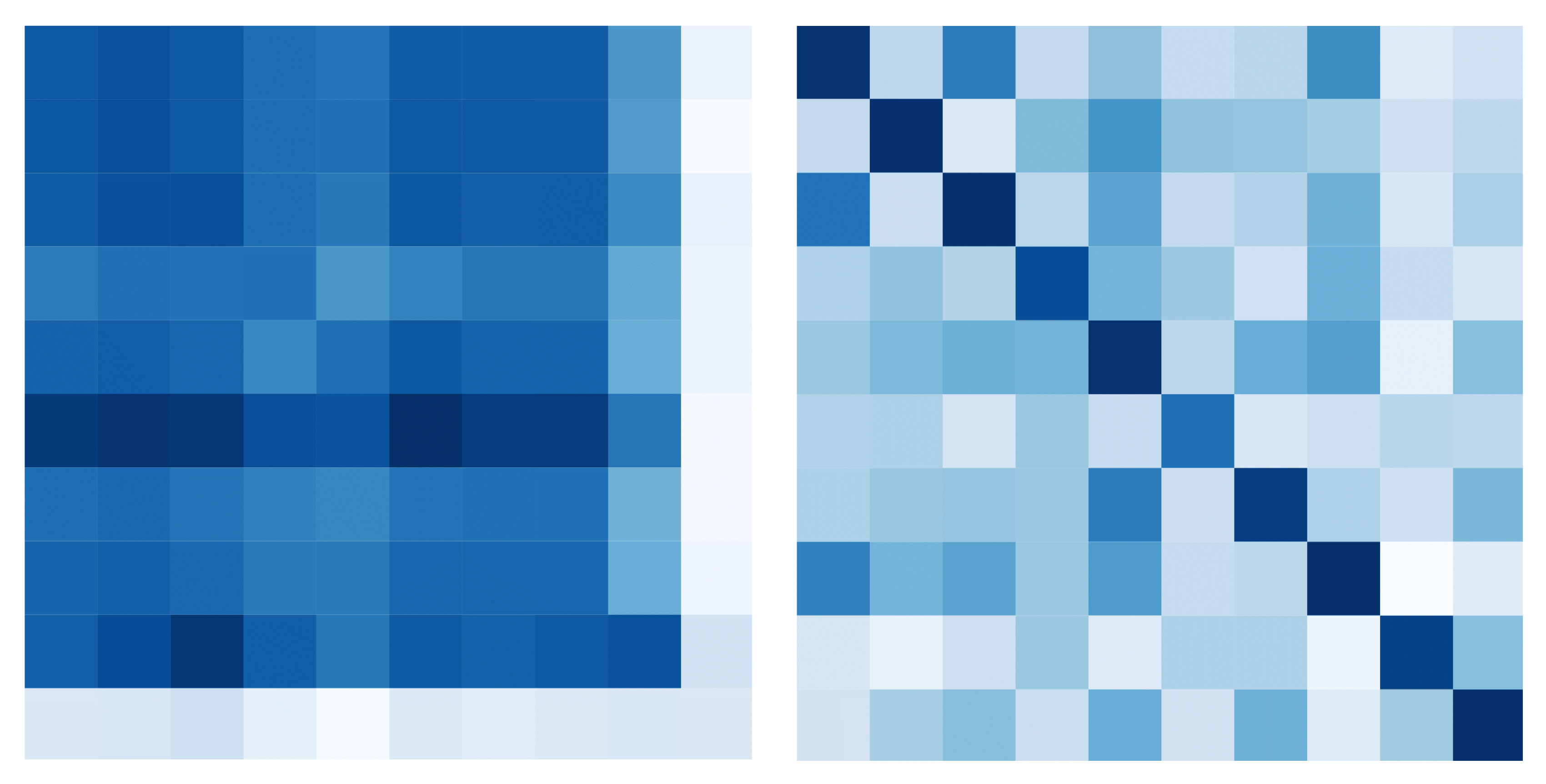}}
     \subfloat[STL-10 \label{fig:Senu04_5}]{\includegraphics[width=0.3\linewidth]{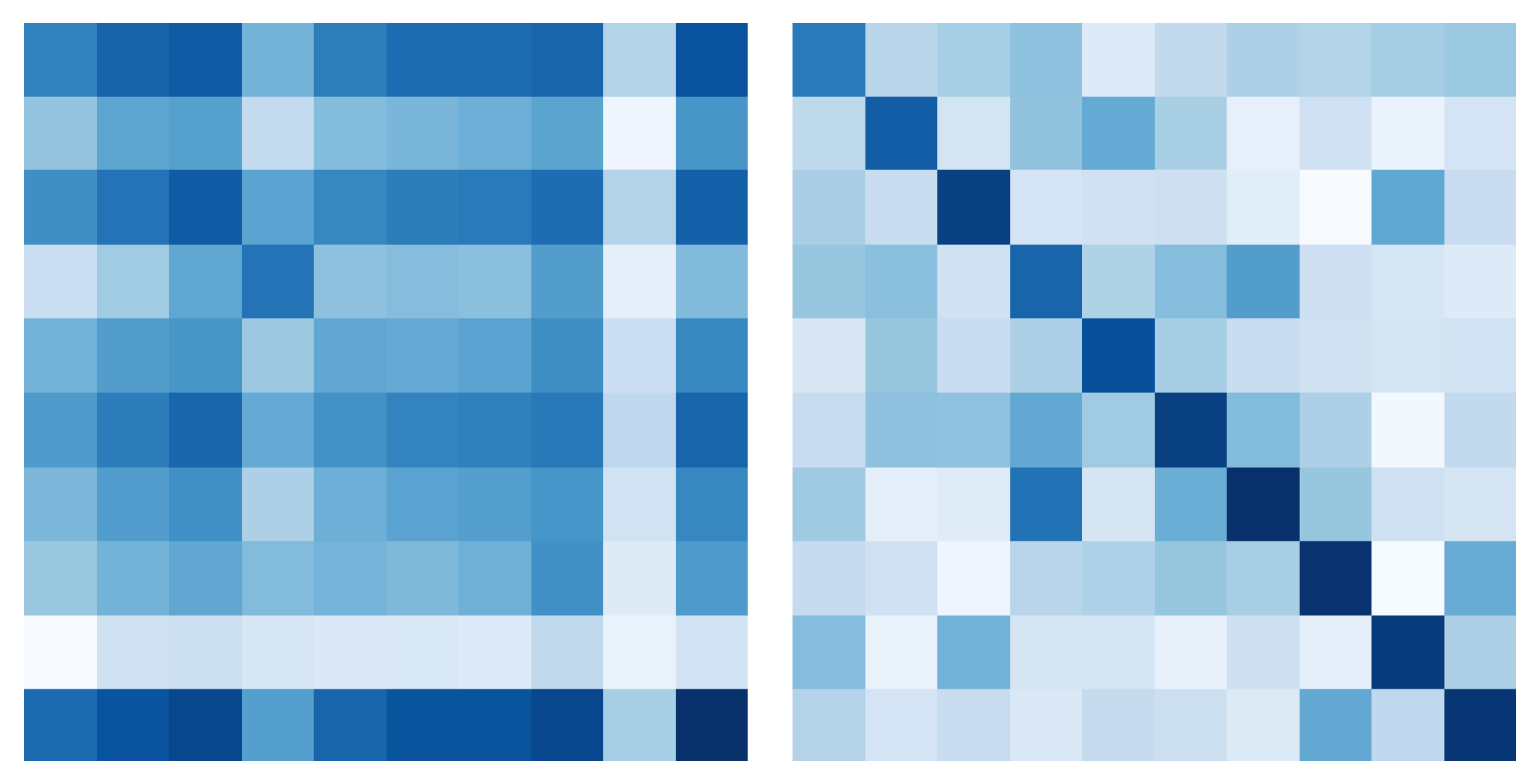}}
      \subfloat[Flower-17 \label{fig:flower}]{\includegraphics[width=0.3\linewidth]{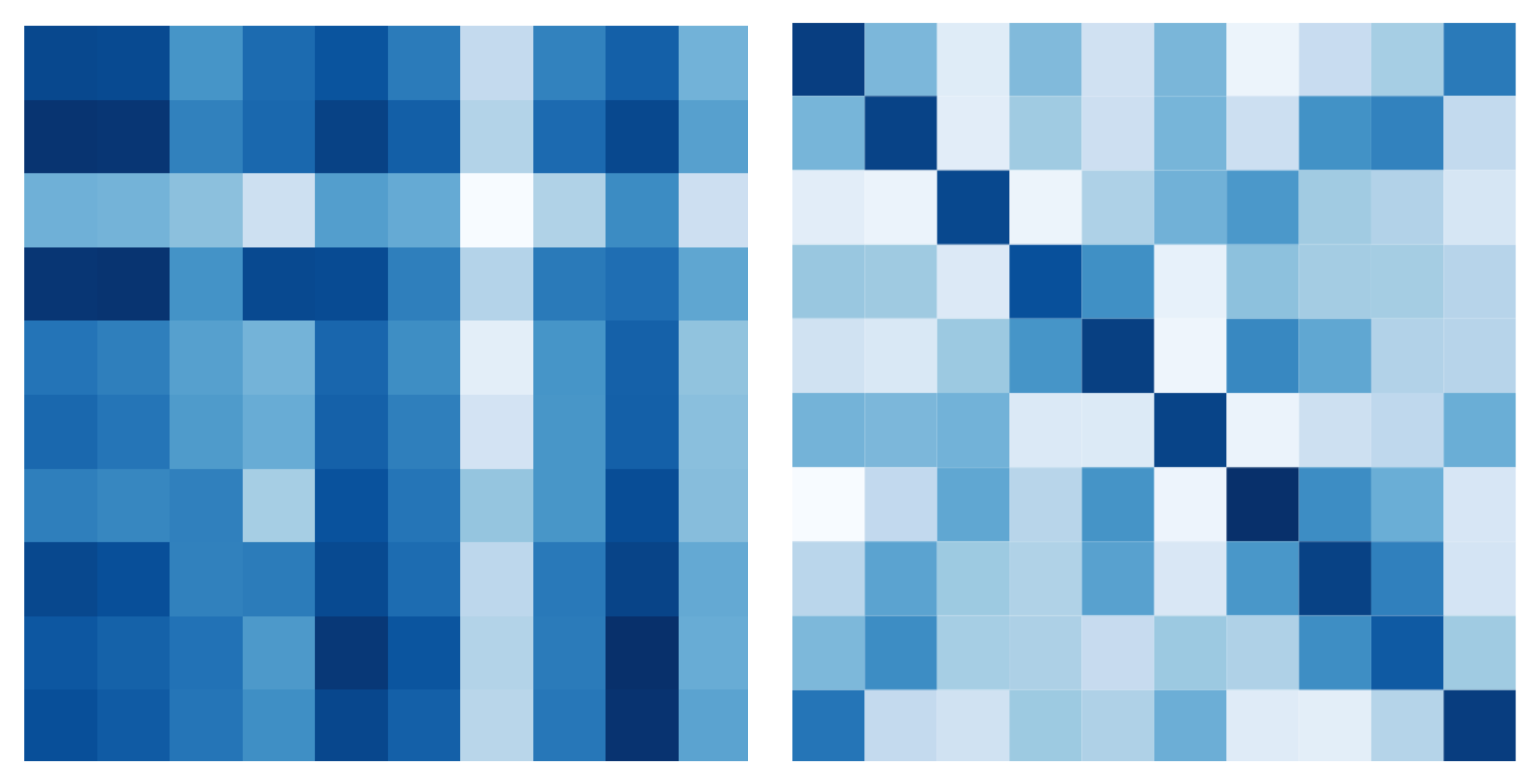}}
    \caption{The cross-correlation among latent representations: Each matrix, sized $10\times10$, illustrates the pairwise cross-correlations between encoded representations of the datasets. Darker diagonal entries, tending towards perfect correlation, signify better performance in terms of both lossy compression and preservation of mutual information.} 
    \label{fig:Senu04}
\end{figure*}

Fig. \ref{fig:Senu04} shows the cross-correlation among latent representations of the proposed semantic encoders, illustrating their effectiveness in achieving redundancy reduction and improved invariant representation. The figure compares the cross-correlation matrices of the datasets before and after training. Notably, these matrices tend towards a diagonal structure, with the off-diagonal terms approaching $0$. This convergence signifies a substantial reduction in redundancy among the encoded features, aligning with the ideal scenario where features capture mutual invariant aspects of the data.

\begin{figure}[t]
\centering
     \subfloat[Cosine similarity\label{fig:TL_cossim}]{\includegraphics[width=0.5\linewidth, height=5.5cm]{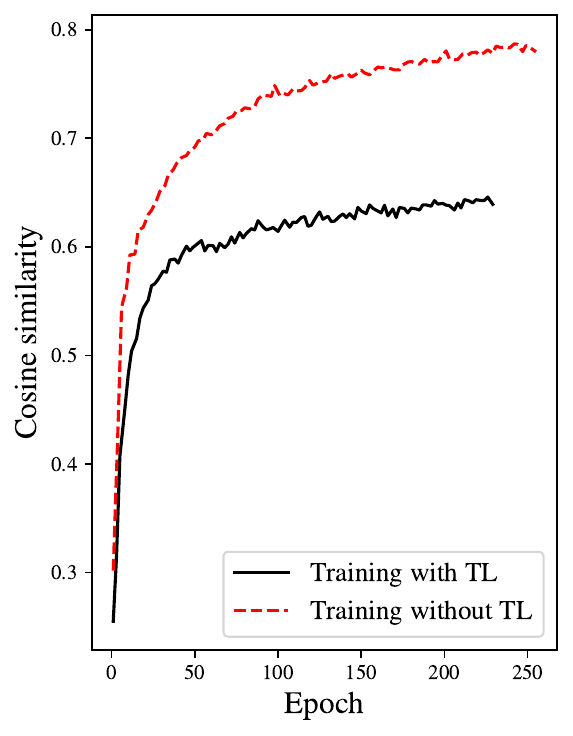}}
     \subfloat[The total loss\label{fig:TL_loss}]{\includegraphics[width=0.5\linewidth, height=5.5cm]{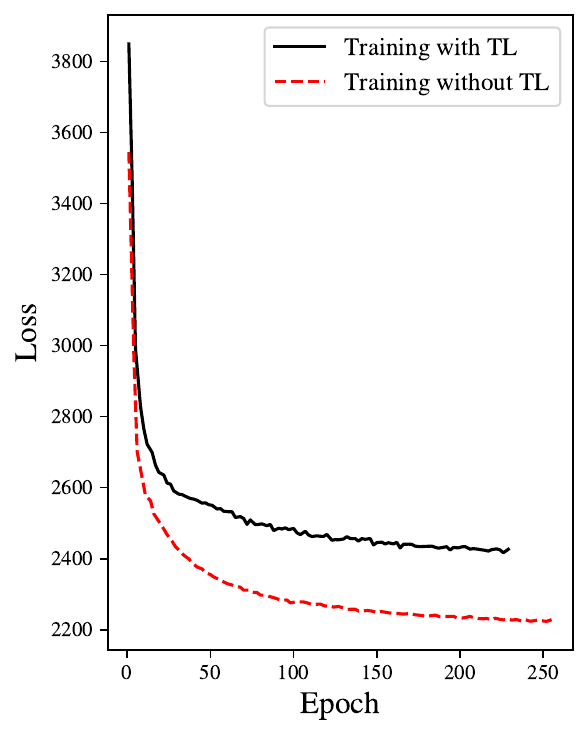}}
    \caption{Transfer learning (TL)-aided SC-GIR with different background knowledge: (a) Cosine similarity versus the number of training epochs, and (b) Total loss versus the number of training epochs.}\label{fig:TL_knowledge}
\end{figure}

The performance of the SC-GIR method, when trained with and without transfer learning (TL), is shown in Fig.~\ref{fig:TL_knowledge}. As can be seen from Fig.~\ref{fig:TL_cossim}, a higher cosine similarity is obtained for SC-GIR trained without TL, indicating superior semantic feature extraction capability. This enhanced extraction suggests that the specialized semantic feature extractor can adapt and tailor its representations more effectively to new communication environments without relying on TL. Additionally, Fig.~\ref{fig:TL_loss} shows the total loss over training epochs, revealing a quicker decrease in the loss for the TL-absent SC-GIR method, confirming faster convergence and potentially more efficient training. These results underscore the advantages of our approach in training SC-GIR systems autonomously, without relying on background knowledge transferred through TL. This can be attributed to the fact that the design of the semantic feature extractor and loss function is optimized specifically for this purpose.

\subsection{GOAI Evaluation}

\begin{figure}[t]
\centering
      \subfloat[Impact of the learning rate\label{fig:learning rate}]
     {\includegraphics[width=1\linewidth, height=4cm]{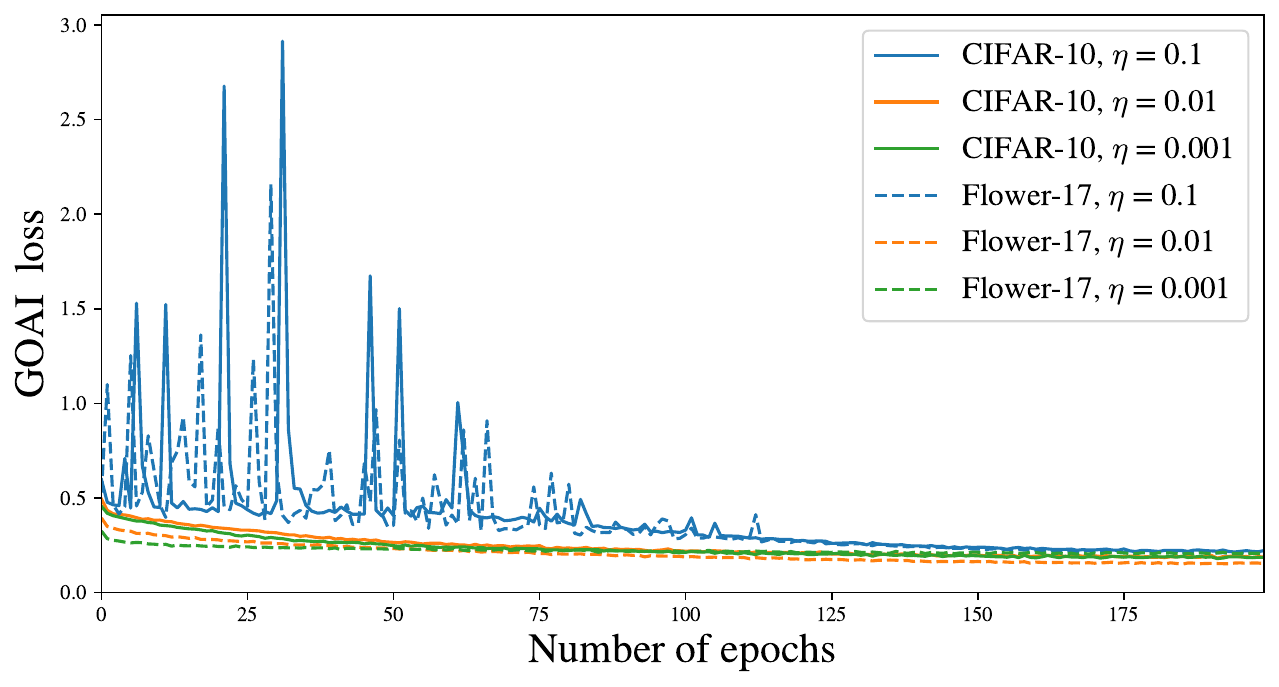}}\\
     \subfloat[Impact of the batch size\label{fig:batch size}]
       {\includegraphics[width=1\linewidth,height=4cm]{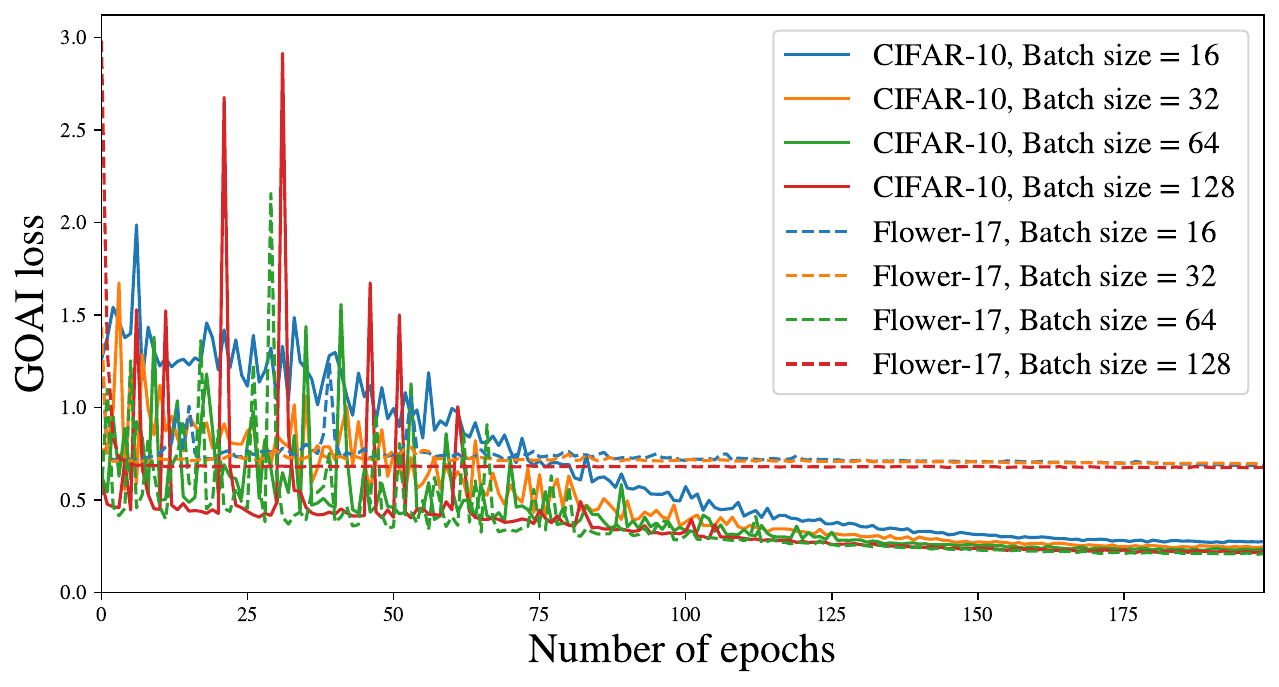}}
    \caption{The GOAI loss on CIFAR-10 \& Flower-17: (a) Impact of the learning rate (batch size of $128$), and (b) Impact of the batch size (learning rate of $0.1$)}
\end{figure}

Fig.~\ref{fig:learning rate} investigates the impact of the learning rate on the training convergence behavior of the SC-GIR model for the downstream goal-oriented AI task across CIFAR-10 and Flower-17 datasets over 200 epochs with a fixed batch size of 128. The results demonstrate that learning rates significantly influence the speed and stability of model convergence for both datasets. Specifically, learning rates of $10^{-2}$ and $10^{-3}$ exhibit smoother and faster convergence. Conversely, a learning rate of $10^{-1}$ introduces volatility in loss, leading to fluctuations during training, while a rate of $10^{-3}$ demonstrates steady convergence. These findings emphasize that a learning rate greater than $0.1$ is essential for faster convergence, thus highlighting the critical role of selecting an appropriate learning rate to enhance the efficiency of the SC-GIR model in capturing task-relevant features and optimizing the training time.

In Fig.~\ref{fig:batch size}, we illustrate the effect of different batch sizes on the training loss of the GOAI task over $200$ epochs with a learning rate of $0.1$. Notably, the training process with a batch size of 128 demonstrates a more stable convergence pattern for CIFAR-10, while the same stability occurs at a batch size of 64 for Flower-17 compared to other smaller batch sizes, evident from the smoother descent in loss value. This observation implies that challenges commonly associated with smaller batch sizes, such as noisy and inconsistent updates, may be alleviated by utilizing larger batch sizes. Additionally, employing a larger batch size has the potential to reduce the overall number of updates required, thereby possibly decreasing the computational expense associated with the GOAI task. Future research could explore optimal batch size selection strategies for similar tasks or domains, considering computational resources and convergence stability.

\begin{table}[t]
\centering
\caption{Accuracy of GOAI for Different Levels of the Noise Power in Eq.~\eqref{eq:wirelesschannel}}
\label{tab:channelvariations}
\resizebox{\columnwidth}{!}{
\begin{tabular}{c||c|c|c|l}
 
\hline
\multirow{2}{*}{$\sigma_n^2$} & \multicolumn{4}{c}{Datasets}                               \\ \cline{2-5} 
                          & CIFAR-10 & STL-10 & CIFAR-100 & \multicolumn{1}{c}{FMNIST} \\ \hline \hline
0.1                       & 73.75    & 75.78  & 71.09     & 68.75                      \\
0.01                      & 74.53    & 77.34  & 77.78     & 72.65                      \\
0.001                     & 76.09    & 78.91  & 78.12     & 73.43             \\
\hline
\end{tabular}
}
\end{table}
\begin{table}[t]
\centering
\caption{F1-scores in Percentages of SC-GIR and DeepJSCC in the Goal-oriented AI Task Under SNR = 25dB.}
\label{tab:F1score}
\resizebox{0.8\columnwidth}{!}{
\begin{tabular}{c||l|l} 
\hline
Dataset   & SC-GIR & DeepJSCC  \\ 
\hline \hline \rowcolor{lavender}
CIFAR-10  &    0.9687   &     0.6655        \\ 
\hline
CIFAR-100 &    0.7968    &      0.5055       \\ 
\hline \rowcolor{lavender}
EMNIST   &    0.9220    &     0.8555        \\ 
\hline
FMNIST   &    0.8654    &     0.7762        \\ 
\hline \rowcolor{lavender}
Flower-17 &     0.9843   &     0.7056        \\ 
\hline
STL-10    &   0.7578     &     0.5625          \\
\hline
\end{tabular}
}
\end{table}

\begin{table*}[t]
\centering
\caption{IoU† Segmentation Performance Comparison on Cityscapes Dataset}
\label{tab:iou_comparison_part1}
% \adjustbox{}{

\begin{tabular}{l||cccccccccc}
\hline
\textbf{Method} & \textbf{mean} & \textbf{road} & \textbf{sidew.} & \textbf{build.} & \textbf{wall} & \textbf{fence} & \textbf{pole} & \textbf{light} & \textbf{sign} & \textbf{veget} \\
\hline \hline
SegFormer$_{512}$ \cite{xie2021segformer} & 41.5 & 68.1 & 33.5 & 62.0 & 24.1 & 19.5 & 29.0 & 36.8 & 38.5 & 60.3 \\
\hline
Refign \cite{bruggemann2023refign} & 57.8 & 82.3 & 54.0 & 80.1 & 35.1 & 25.0 & 41.5 & 53.6 & 51.9 & 80.5 \\
MGCDA \cite{sakaridis2020map} & 39.2 & 67.5 & 22.1 & 63.8 & 13.0 & 16.8 & 26.1 & 41.5 & 42.0 & 65.0 \\
\hline
\textbf{SC-GIR (ours)} & \textbf{63.5} & \textbf{88.1} & \textbf{67.2} & \textbf{83.5} & \textbf{43.0} & \textbf{35.5} & \textbf{45.3} & \textbf{59.8} & \textbf{60.1} & \textbf{82.4} \\
\hline

 & \textbf{terrain} & \textbf{sky} & \textbf{person} & \textbf{rider} & \textbf{car} & \textbf{truck} & \textbf{bus} & \textbf{train} & \textbf{motorc.} & \textbf{bicycle} \\
\hline
SegFormer$_{512}$ \cite{xie2021segformer} & 28.2 & 71.4 & 29.6 & 14.1 & 69.8 & 40.1 & 51.3 & 23.8 & 32.7 & 40.2 \\
\hline
Refign \cite{bruggemann2023refign} & 50.1 & 92.6 & 52.3 & 28.5 & 78.0 & 57.9 & 63.0 & 36.8 & 40.1 & 46.5 \\
MGCDA \cite{sakaridis2020map} & 24.5 & 80.1 & 38.4 & 16.2 & 68.1 & 34.0 & 35.7 & 22.5 & 29.1 & 37.3 \\
\hline
\end{tabular}
% }
\end{table*}

Table~\ref{tab:channelvariations} demonstrates the robustness of the latent representations derived from the semantic encoder when subjected to the noise power. The table showcases the impact of noise variance on the performance of the GOAI task across different datasets, varying in complexity from those with fewer targets to those with more targets. Notably, even at a relatively high noise variance $(\sigma^2 = 0.1)$, all datasets maintain a performance level exceeding $70\%$, showing the resilience of the latent representations against noise. Moreover, there is a clear trend of increasing accuracy as the noise variance decreases from $0.1$ to $0.001$, indicating the effectiveness of the semantic encoding in preserving information critical for the GOAI task under varying noise conditions. This resilience holds promise for applications in wireless communication systems, where robustness to noise is crucial for maintaining reliable performance. 

\begin{table}[t]
\centering
\caption{Leave-one-domain-out results on PACS with ResNet-34}
\label{tab:pacs_semantic_comm}
\begin{tabular}{l||cccc|c}
\hline
Methods & Art & Cartoon & Photo & Sketch & Avg. \\
\hline\hline
SemCC & 67.83 & 62.47 & 87.92 & 58.31 & 69.13 \\
SemRE & 70.24 & 64.85 & 89.16 & 59.72 & 70.99 \\
DeepJSCC & 62.19 & 57.34 & 83.61 & 52.78 & 63.98 \\
DeepSC & 64.52 & 59.83 & 84.27 & 54.95 & 65.89 \\
\hline
SC-GIR & \textbf{76.89} & \textbf{72.32} & \textbf{96.54} & 58.70 & \textbf{76.11}\\
\hline
\end{tabular}
\end{table}

F1-score, ranging from $0$ to $1$, is an important metric for evaluating learned latent representations in GOAI tasks. It balances precision and recall, which are crucial for accurately identifying relevant instances while minimizing false positives.
In Table~\ref{tab:F1score}, the F1 scores attained by the proposed SC-GIR method in comparison to the DeepJSCC baseline are provided. These assessments are conducted at an SNR of $25$ dB, assessing the models' resilience in noisy environments. SC-GIR consistently outperforms DeepJSCC across all datasets examined. Specifically, SC-GIR achieves F1 scores ranging from $0.7578$ on the STL-10 dataset to 0.9843 on Flower-17, highlighting its capability to extract task-relevant information from latent representations. In contrast, DeepJSCC's performance is comparatively suboptimal, with F1 scores spanning from $0.5055$ (on CIFAR-100) to $0.8555$ (on EMNIST). The substantial improvement in F1 scores observed with SC-GIR underscores the efficacy of its tailored approach in minimizing redundancy and capturing informative features in the latent space, thereby enhancing its performance in goal-oriented AI tasks.

\subsection{Generalization Capability}

The core idea behind SC-GIR is to learn dense and invariant semantic representations by minimizing redundancy through a cross-correlation loss. Unlike classification, which operates at the image level, semantic segmentation is a dense and pixel-level task. We hypothesize that a model learning truly general-purpose semantic features should excel at both, as our representations aim to capture the full semantic structure of an image, not just identify a single class. To demonstrate this generality, we extend SC-GIR to semantic segmentation and domain generalization. For segmentation, we train SC-GIR's semantic encoder with a standard decoder head and evaluate it on the challenging Cityscapes dataset using mean Intersection-over-Union (mIoU).

As shown in Table~\ref{tab:iou_comparison_part1}, SC-GIR demonstrates strong semantic segmentation performance on the Cityscapes dataset, outperforming established baselines. It achieves a mean IoU of $63.5$, marking a $5.7$-point improvement over the strong Refign baseline \cite{bruggemann2023refign}. This advantage spans nearly all categories, with especially large gains on structured objects such as \texttt{train} $(+38.6)$ \texttt{bus} $(+13.3)$ and \texttt{wall} $(+7.9)$. While Refign remains competitive on some smaller classes like \texttt{rider} and \texttt{bicycle}, our model's significant overall improvement provides strong evidence that its learned invariant representations go beyond classification. They effectively capture the rich, dense semantics necessary for comprehensive scene understanding.

To validate generalization empirically, we conduct a leave-one-domain-out experiment on the PACS dataset \cite{li2017deeper}, a standard benchmark for domain generalization. The considered models are trained on three of the four domains (Photo, Art, Cartoon, Sketch) and tested on the held-out domain. This procedure is repeated for all domain splits, and average accuracy is used to measure generalization capability. As shown in Table~\ref{tab:pacs_semantic_comm}, SC-GIR achieves an average accuracy of $76.11\%$, outperforming SemRE by over $5\%$ and DeepJSCC by more than $12\%$. This strong performance on unseen domains highlights SC-GIR's scalability. By learning invariant semantic features, it avoids overfitting to source domains and generalizes effectively, reducing the need for retraining in visually distinct environments.

\section{Conclusion}\label{sec_Conclusion}
This paper presented SC-GIR, a novel framework for goal-oriented semantic communication in IoT networks, which leverages self-supervised learning to derive invariant and task-relevant feature representations. By optimizing the cross-correlation matrix between multiple augmented views of the data, SC-GIR effectively filters out redundant information and preserves semantic features critical to downstream tasks. This allows for compact, bandwidth-efficient encoding, especially suited for resource-constrained communication environments. We conducted comprehensive evaluations on CIFAR-10, STL-10, and Flower-17 datasets, demonstrating that SC-GIR consistently outperforms state-of-the-art methods such as DeepJSCC and SEM-RE. Specifically, SC-GIR achieves over $85\%$ classification accuracy under AWGN channels and exceeds $80\%$ under Rayleigh fading at a compression ratio of $0.1$, highlighting its robustness under both noise and severe bandwidth constraints. A key advantage of SC-GIR lies in its use of self-supervised learning, which reduces reliance on labeled data. This enables greater adaptability to diverse tasks, deployment settings, and channel conditions, facilitating scalability across heterogeneous IoT ecosystems. Additionally, SC-GIR achieves strong noise resilience and semantic fidelity across a wide range of SNR levels while significantly minimizing communication overhead.

Beyond image-based tasks, the design principles of SC-GIR generalize to other data modalities such as text and audio, making it a versatile and extensible solution for future semantic communication systems. By addressing critical challenges such as data scarcity, bandwidth limitations, and semantic integrity, SC-GIR represents a meaningful step toward intelligent, robust, and efficient inter-device communication in next-generation IoT and edge computing networks.

{
% \balance
\bibliographystyle{IEEEtran}
\bibliography{IEEE}

\begin{thebibliography}{10}
\providecommand{\url}[1]{#1}
\csname url@samestyle\endcsname
\providecommand{\newblock}{\relax}
\providecommand{\bibinfo}[2]{#2}
\providecommand{\BIBentrySTDinterwordspacing}{\spaceskip=0pt\relax}
\providecommand{\BIBentryALTinterwordstretchfactor}{4}
\providecommand{\BIBentryALTinterwordspacing}{\spaceskip=\fontdimen2\font plus
\BIBentryALTinterwordstretchfactor\fontdimen3\font minus \fontdimen4\font\relax}
\providecommand{\BIBforeignlanguage}[2]{{%
\expandafter\ifx\csname l@#1\endcsname\relax
\typeout{** WARNING: IEEEtran.bst: No hyphenation pattern has been}%
\typeout{** loaded for the language `#1'. Using the pattern for}%
\typeout{** the default language instead.}%
\else
\language=\csname l@#1\endcsname
\fi
#2}}
\providecommand{\BIBdecl}{\relax}
\BIBdecl

\bibitem{2024-SemCom-Survey2}
T.-H. Vu, S.~K. Jagatheesaperumal, M.-D. Nguyen, N.~V. Huynh, S.~Kim, and Q.-V. Pham, ``{Applications of Generative AI (GAI) for Mobile and Wireless Networking: A Survey},'' \emph{IEEE Internet of Things Journ.}, Aug. 2024.

\bibitem{cabrera20216g}
J.~A. Cabrera, H.~Boche, C.~Deppe, R.~F. Schaefer, C.~Scheunert, and F.~H.~P. Fitzek, \emph{6{G} and the post‐Shannon theory}.\hskip 1em plus 0.5em minus 0.4em\relax John Wiley \& Sons, Ltd, 2022, pp. 271--294.

\bibitem{ZhijinProc24}
Z.~Qin, L.~Liang, Z.~Wang, S.~Jin, X.~Tao, W.~Tong, and G.~Y. Li, ``{AI} empowered wireless communications: From bits to semantics,'' \emph{Proc. IEEE}, vol. 112, no.~7, pp. 621--652, 2024.

\bibitem{9994683}
A.~Mostaani, T.~X. Vu, S.~K. Sharma, V.-D. Nguyen, Q.~Liao, and S.~Chatzinotas, ``Task-oriented communication design in cyber-physical systems: A survey on theory and applications,'' \emph{IEEE Access}, vol.~10, pp. 133\,842--133\,868, 2022.

\bibitem{hu2022robust}
Q.~Hu, G.~Zhang, Z.~Qin, Y.~Cai, G.~Yu, and G.~Y. Li, ``Robust semantic communications against semantic noise,'' in \emph{2022 IEEE 96th Vehicular Technology Conference (VTC2022-Fall)}, 2022.

\bibitem{shao2021learning}
J.~Shao, Y.~Mao, and J.~Zhang, ``Learning task-oriented communication for edge inference: An information bottleneck approach,'' \emph{IEEE Journal on Selected Areas in Communications}, vol.~40, no.~1, pp. 197--211, 2021.

\bibitem{10458014}
Y.~Wang, S.~Guo, Y.~Deng, H.~Zhang, and Y.~Fang, ``Privacy-preserving task-oriented semantic communications against model inversion attacks,'' \emph{IEEE Transactions on Wireless Communications}, vol.~23, no.~8, pp. 10\,150--10\,165, 2024.

\bibitem{10483549}
Y.~Lin, Z.~Gao, H.~Du, D.~Niyato, J.~Kang, Z.~Xiong, and Z.~Zheng, ``Blockchain-based efficient and trustworthy aigc services in metaverse,'' \emph{IEEE Transactions on Services Computing}, Mar. 2024.

\bibitem{9252948}
H.~Xie and Z.~Qin, ``A lite distributed semantic communication system for internet of things,'' \emph{IEEE Journal on Selected Areas in Communications}, vol.~39, no.~1, pp. 142--153, 2021.

\bibitem{bourtsoulatze2019deep}
E.~Bourtsoulatze, D.~Burth~Kurka, and D.~Gündüz, ``Deep joint source-channel coding for wireless image transmission,'' \emph{IEEE Trans. Cogn. Commun. and Network.}, vol.~5, no.~3, pp. 567--579, 2019.

\bibitem{liu2021self}
X.~Liu, F.~Zhang, Z.~Hou, L.~Mian, Z.~Wang, J.~Zhang, and J.~Tang, ``Self-supervised learning: Generative or contrastive,'' \emph{IEEE Transactions on Knowledge and Data Engineering}, vol.~35, no.~1, pp. 857--876, 2023.

\bibitem{shannon1948mathematical}
C.~E. Shannon, ``A mathematical theory of communication,'' \emph{The Bell System Technical Journal}, vol.~27, no.~3, pp. 379--423, 1948.

\bibitem{jerri1977shannon}
A.~Jerri, ``The {Shannon} sampling theorem—{I}ts various extensions and applications: A tutorial review,'' \emph{Proc. IEEE}, vol.~65, no.~11, pp. 1565--1596, 1977.

\bibitem{shi2021semantic}
G.~Shi, Y.~Xiao, Y.~Li, and X.~Xie, ``From semantic communication to semantic-aware networking: Model, architecture, and open problems,'' \emph{IEEE Communications Magazine}, vol.~59, no.~8, pp. 44--50, 2021.

\bibitem{al2017internet}
S.~Al-Sarawi, M.~Anbar, K.~Alieyan, and M.~Alzubaidi, ``Internet of {T}hings ({IoT}) communication protocols: Review,'' in \emph{Proc. 8th Inter. Confe. Infor. Tech.}\hskip 1em plus 0.5em minus 0.4em\relax IEEE, 2017, pp. 685--690.

\bibitem{bayilmics2022survey}
C.~Bayılmış, M.~A. Ebleme, Ünal Çavuşoğlu, K.~Küçük, and A.~Sevin, ``A survey on communication protocols and performance evaluations for internet of things,'' \emph{Digi. Commun. and Net.}, vol.~8, no.~6, pp. 1094--1104, 2022.

\bibitem{stankovic2014research}
J.~A. Stankovic, ``Research directions for the {Internet of Things},'' \emph{IEEE Internet of Things J.}, vol.~1, no.~1, pp. 3--9, 2014.

\bibitem{2021-SEM-DeepSC}
H.~Xie, Z.~Qin, G.~Y. Li, and B.-H. Juang, ``Deep learning enabled semantic communication systems,'' \emph{IEEE Trans. Sign. Process.}, Sep. 2021.

\bibitem{2023-SemCom-UDeepSC}
G.~Zhang, Q.~Hu, Z.~Qin, Y.~Cai, G.~Yu, and X.~Tao, ``{A Unified Multi-Task Semantic Communication System for Multimodal Data},'' \emph{arXiv preprint arXiv:2209.07689}, Aug. 2023.

\bibitem{2022-SemCom-MUDeepSC}
H.~Xie, Z.~Qin, X.~Tao, and K.~B. Letaief, ``{Task-Oriented Multi-User Semantic Communications},'' \emph{IEEE Jour. of Sel. Areas in Comm.}, Mar. 2022.

\bibitem{2023-SemCom-MemDeepSC}
H.~Xie, Z.~Qin, and G.~Y. Li, ``Semantic communication with memory,'' \emph{IEEE Jour. of Sel. Areas in Comm.}, Jun. 2023.

\bibitem{2022-SEM-AdaptableSemanticCompression}
C.~Liu, C.~Guo, Y.~Yang, and N.~Jiang, ``{Adaptable Semantic Compression and Resource Allocation for Task-Oriented Communications},'' \emph{IEEE Trans. Cogn. Comm. and Netw.}, Jun. 2024.

\bibitem{2020-SemCom-DJSCCF}
D.~B. Kurka and D.~Gündüz, ``{DeepJSCC-f: Deep Joint Source-Channel Coding of Images With Feedback},'' \emph{IEEE Jour. of Sel. Areas in Comm.}, May 2020.

\bibitem{2019-SemCOm-DJSCC-WIT}
E.~Bourtsoulatze, D.~Burth~Kurka, and D.~Gündüz, ``{Deep Joint Source-Channel Coding for Wireless Image Transmission},'' \emph{IEEE Trans. Cogn. Comm. and Netw.}, Apr. 2019.

\bibitem{2024-SemCom-AdaSem}
Q.~Liao and T.-Y. Tung, ``{AdaSem: Adaptive Goal-Oriented Semantic Communications for End-to-End Camera Relocalization},'' in \emph{INFOCOM}, May 2024.

\bibitem{2024-SemCom-SemCC}
S.~Tang, Q.~Yang, L.~Fan, X.~Lei, A.~Nallanathan, and G.~K. Karagiannidis, ``Contrastive learning-based semantic communications,'' \emph{IEEE Trans. on Comm.}, Oct. 2024.

\bibitem{2024-SemCom-DeepMA}
W.~Zhang, K.~Bai, S.~Zeadally, H.~Zhang, H.~Shao, H.~Ma, and V.~C.~M. Leung, ``{DeepMA}: End-to-end deep multiple access for wireless image transmission in semantic communication,'' \emph{IEEE Trans. Cogn. Comm. and Netw.}, Apr. 2024.

\bibitem{2023-SemCom-GenerativeJSCC}
E.~Erdemir, T.-Y. Tung, P.~L. Dragotti, and D.~Gündüz, ``Generative joint source-channel coding for semantic image transmission,'' \emph{IEEE Jour. of Sel. Areas in Comm.}, Aug. 2023.

\bibitem{radford21a}
\BIBentryALTinterwordspacing
A.~Radford, J.~W. Kim, C.~Hallacy, A.~Ramesh, G.~Goh, S.~Agarwal, G.~Sastry, A.~Askell, P.~Mishkin, J.~Clark, G.~Krueger, and I.~Sutskever, ``Learning transferable visual models from natural language supervision,'' in \emph{Proceedings of the 38th International Conference on Machine Learning}, ser. Proceedings of Machine Learning Research, M.~Meila and T.~Zhang, Eds., vol. 139.\hskip 1em plus 0.5em minus 0.4em\relax PMLR, 18--24 Jul 2021, pp. 8748--8763. [Online]. Available: \url{https://proceedings.mlr.press/v139/radford21a.html}
\BIBentrySTDinterwordspacing

\bibitem{lopez2018information}
R.~Lopez, J.~Regier, M.~I. Jordan, and N.~Yosef, ``Information constraints on auto-encoding variational bayes,'' \emph{Advances in neural information processing systems}, vol.~31, 2018.

\bibitem{xie2021deep}
H.~Xie, Z.~Qin, G.~Y. Li, and B.-H. Juang, ``Deep learning enabled semantic communication systems,'' \emph{IEEE Trans. Signal Process.}, vol.~69, pp. 2663--2675, 2021.

\bibitem{federici2020learning}
M.~Federici, A.~Dutta, P.~Forré, N.~Kushman, and Z.~Akata, ``Learning robust representations via multi-view information bottleneck,'' in \emph{International Conference on Learning Representations}, 2020.

\bibitem{shwartz2024compress}
R.~Shwartz~Ziv and Y.~LeCun, ``To compress or not to compress—self-supervised learning and information theory: A review,'' \emph{Entropy}, vol.~26, no.~3, p. 252, 2024.

\bibitem{tishby2000information}
N.~Tishby and N.~Zaslavsky, ``{Deep learning and the information bottleneck principle},'' in \emph{2015 IEEE Information Theory Workshop (ITW)}, 2015, pp. 1--5.

\bibitem{10436784}
H.~Li, W.~Yu, H.~He, J.~Shao, S.~Song, J.~Zhang, and K.~B. Letaief, ``Task-oriented communication with out-of-distribution detection: An information bottleneck framework,'' in \emph{GLOBECOM 2023 - 2023 IEEE Global Communications Conference}, Feb. 2024.

\bibitem{2022-IL-DIR}
Y.~Wu, X.~Wang, A.~Zhang, X.~He, and T.-S. Chua, ``{Discovering Invariant Rationales for Graph Neural Networks},'' in \emph{Int. Conf. Learn. Represent.}, May 2022.

\bibitem{2021-IL-SSDA}
B.~Li, Y.~Wang, S.~Zhang, D.~Li, K.~Keutzer, T.~Darrell, and H.~Zhao, ``Learning invariant representations and risks for semi-supervised domain adaptation,'' in \emph{IEEE Conf. Comput. Vis. Pattern Recog.}, Jun. 2021, pp. 1104--1113.

\bibitem{wu2021rethinking}
C.~Wu, F.~Wu, and Y.~Huang, ``Rethinking infonce: How many negative samples do you need?'' in \emph{Proc. 31st Inter. Joint Conf. Artificial Intelligence, {IJCAI-22}}, 7 2022, pp. 2509--2515.

\bibitem{bardes2021vicreg}
A.~Bardes, J.~Ponce, and Y.~LeCun, ``{VICR}eg: Variance-invariance-covariance regularization for self-supervised learning,'' in \emph{International Conference on Learning Representations}, 2022.

\bibitem{zbontar2021barlow}
J.~Zbontar, L.~Jing, I.~Misra, Y.~LeCun, and S.~Deny, ``Barlow twins: Self-supervised learning via redundancy reduction,'' in \emph{Proc. 38th Inter. Conf. Machine Learning, (ICML)}, vol. 139, 2021, pp. 12\,310--12\,320.

\bibitem{2022-DG-CIRL}
F.~Lv, J.~Liang, S.~Li, B.~Zang, C.~H. Liu, Z.~Wang, and D.~Liu, ``Causality inspired representation learning for domain generalization,'' in \emph{IEEE Conf. Comput. Vis. Pattern Recog.}, June 2022, pp. 8046--8056.

\bibitem{2024-FL-FCCL}
W.~Huang \emph{et~al.}, ``Generalizable heterogeneous federated cross-correlation and instance similarity learning,'' \emph{IEEE Trans. Patt. Analysis and Machine Intelligence}, vol.~46, no.~2, pp. 712--728, 2024.

\bibitem{chen2020simple}
T.~Chen, S.~Kornblith, M.~Norouzi, and G.~Hinton, ``A simple framework for contrastive learning of visual representations,'' in \emph{Proc. Inter. Conf. Machine Learning}.\hskip 1em plus 0.5em minus 0.4em\relax PMLR, 2020, pp. 1597--1607.

\bibitem{wang2022rethinking}
H.~Wang, X.~Guo, Z.~Deng, and Y.~Lu, ``Rethinking minimal sufficient representation in contrastive learning,'' in \emph{Proc. IEEE/CVF Conf. Comput. Vision and Patt. Recog. (CVPR)}, 2022, pp. 16\,020--16\,029.

\bibitem{tishby2015deep}
N.~Tishby and N.~Zaslavsky, ``{Deep learning and the information bottleneck principle},'' in \emph{IEEE Infor. Theory Workshop (ITW)}, 2015, pp. 1--5.

\bibitem{krizhevsky2009learning}
A.~Krizhevsky, G.~Hinton \emph{et~al.}, ``Learning multiple layers of features from tiny images,'' \emph{Department of Computer Science, University of Toronto}, 2009.

\bibitem{cohen2017emnist}
G.~Cohen, S.~Afshar, J.~Tapson, and A.~van Schaik, ``{EMNIST}: Extending mnist to handwritten letters,'' in \emph{2017 International Joint Conference on Neural Networks (IJCNN)}, 2017, pp. 2921--2926.

\bibitem{xiao2017fashion}
H.~Xiao, K.~Rasul, and R.~Vollgraf, ``Fashion-{MNIST}: A novel image dataset for benchmarking machine learning algorithms,'' \emph{arXiv preprint arXiv:1708.07747}, 2017.

\bibitem{coates2011analysis}
A.~Coates, A.~Ng, and H.~Lee, ``An analysis of single-layer networks in unsupervised feature learning,'' in \emph{Proc. 14th Inter. Conf. Artificial Intelligence and Statistics}, vol.~15.\hskip 1em plus 0.5em minus 0.4em\relax Fort Lauderdale, FL, USA: PMLR, 11--13 Apr 2011, pp. 215--223.

\bibitem{1640927}
M.-E. Nilsback and A.~Zisserman, ``A visual vocabulary for flower classification,'' in \emph{IEEE Conf. Comput. Vision and Patt. Recog.(CVPR'06)}, vol.~2, 2006, pp. 1447--1454.

\bibitem{cordts2016cityscapes}
M.~Cordts, M.~Omran, S.~Ramos, T.~Rehfeld, M.~Enzweiler, R.~Benenson, U.~Franke, S.~Roth, and B.~Schiele, ``The cityscapes dataset for semantic urban scene understanding,'' in \emph{Proceedings of the IEEE conference on computer vision and pattern recognition}, 2016, pp. 3213--3223.

\bibitem{li2017deeper}
D.~Li, Y.~Yang, Y.-Z. Song, and T.~M. Hospedales, ``Deeper, broader and artier domain generalization,'' in \emph{Proceedings of the IEEE international conference on computer vision}, 2017, pp. 5542--5550.

\bibitem{he2016deep}
K.~He, X.~Zhang, S.~Ren, and J.~Sun, ``Deep residual learning for image recognition,'' in \emph{2016 IEEE Conference on Computer Vision and Pattern Recognition (CVPR)}, 2016, pp. 770--778.

\bibitem{hendrycks2016gaussian}
D.~Hendrycks and K.~Gimpel, ``Gaussian error linear units {(GELU)},'' \emph{arXiv preprint arXiv:1606.08415}, 2016.

\bibitem{kingma2014adam}
D.~P. Kingma and J.~Ba, ``Adam: A method for stochastic optimization,'' \emph{CoRR}, vol. abs/1412.6980, 2014.

\bibitem{he2015delving}
K.~He, X.~Zhang, S.~Ren, and J.~Sun, ``Delving deep into rectifiers: Surpassing human-level performance on imagenet classification,'' in \emph{IEEE Inter. Conf. Computer Vision (ICCV)}, 2015, pp. 1026--1034.

\bibitem{loshchilov2016sgdr}
I.~Loshchilov and F.~Hutter, ``{SGDR}: Stochastic gradient descent with warm restarts,'' in \emph{Inter. Conf. Learning Represent.}, 2017.

\bibitem{7780459}
K.~He, X.~Zhang, S.~Ren, and J.~Sun, ``Deep residual learning for image recognition,'' in \emph{2016 IEEE Conference on Computer Vision and Pattern Recognition (CVPR)}, 2016, pp. 770--778.

\bibitem{10530261}
S.~Tang, Q.~Yang, L.~Fan, X.~Lei, A.~Nallanathan, and G.~K. Karagiannidis, ``Contrastive learning-based semantic communications,'' \emph{IEEE Transactions on Communications}, vol.~72, no.~10, pp. 6328--6343, 2024.

\bibitem{9398576}
H.~Xie, Z.~Qin, G.~Y. Li, and B.-H. Juang, ``Deep learning enabled semantic communication systems,'' \emph{IEEE Transactions on Signal Processing}, vol.~69, pp. 2663--2675, 2021.

\bibitem{xie2021segformer}
E.~Xie, W.~Wang, Z.~Yu, A.~Anandkumar, J.~M. Alvarez, and P.~Luo, ``Segformer: Simple and efficient design for semantic segmentation with transformers,'' \emph{Advances in neural information processing systems}, vol.~34, pp. 12\,077--12\,090, 2021.

\bibitem{bruggemann2023refign}
D.~Br{\"u}ggemann, C.~Sakaridis, P.~Truong, and L.~Van~Gool, ``Refign: Align and refine for adaptation of semantic segmentation to adverse conditions,'' in \emph{Proceedings of the IEEE/CVF Winter Conference on Applications of Computer Vision}, 2023, pp. 3174--3184.

\bibitem{sakaridis2020map}
C.~Sakaridis, D.~Dai, and L.~Van~Gool, ``Map-guided curriculum domain adaptation and uncertainty-aware evaluation for semantic nighttime image segmentation,'' \emph{IEEE Transactions on Pattern Analysis and Machine Intelligence}, vol.~44, no.~6, pp. 3139--3153, 2020.

\end{thebibliography}
}

\begin{IEEEbiography}
    [{\includegraphics[width=1.1in,height=1.25in,clip,keepaspectratio]{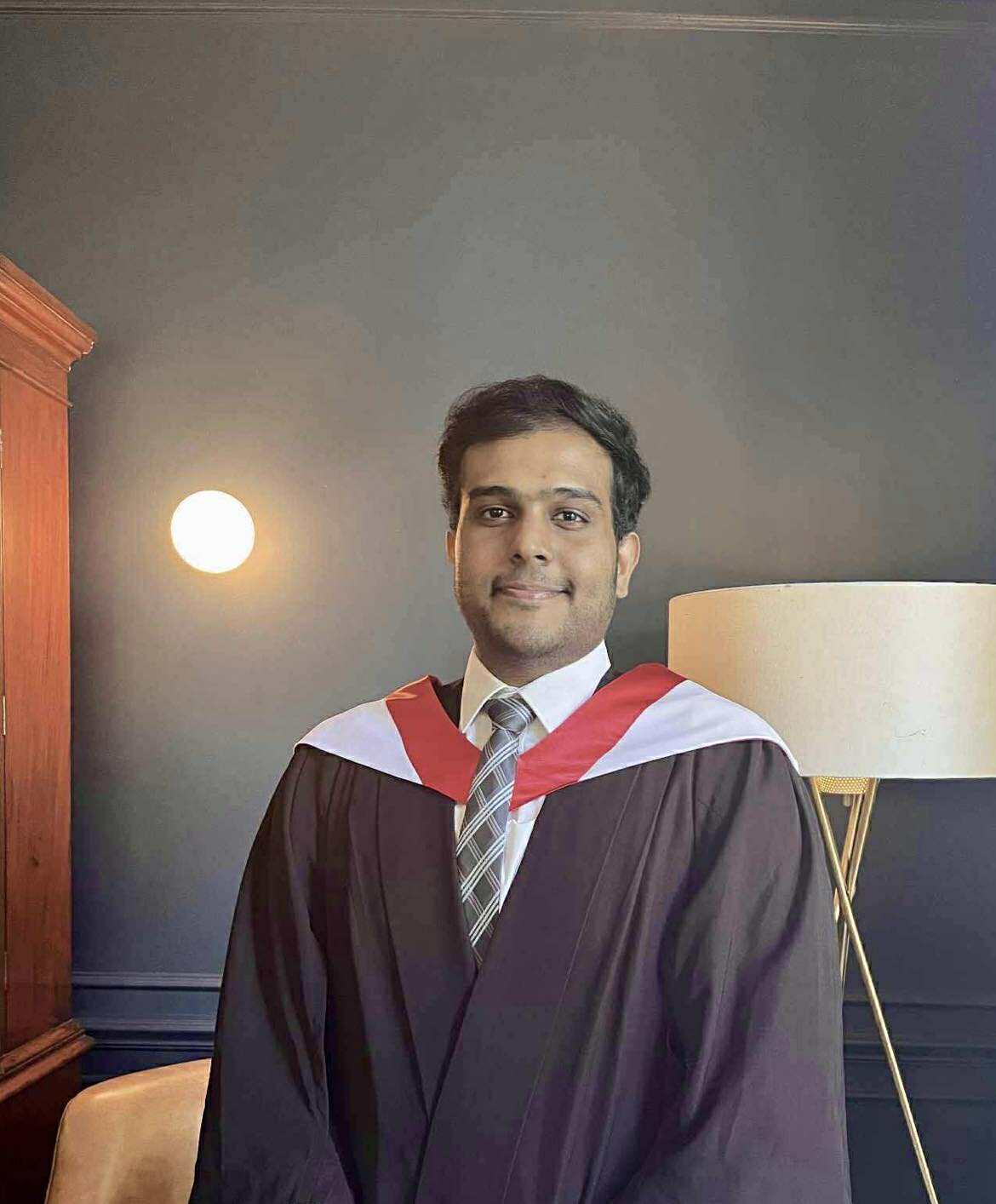}}]{Senura Hansaja Wanasekara}  received the B.S. degree in electronics and telecommunications engineering from the Hanoi University of Science and Technology, Vietnam, in 2023. In the same year, he served as an R\&D engineer at the VNPT R\&D Center, contributing to advanced communication system development. From 2023 to 2024, he was a research assistant at VinUniversity, where he worked on semantic communication systems for next-generation wireless networks. He is currently pursuing a Ph.D. degree at the University of Sydney, Australia, focusing on the intersection of artificial intelligence and computational biology.

    His research interests include generative AI, protein folding, protein language modeling, and semantic communication.

\end{IEEEbiography}

\begin{IEEEbiography}
    [{\includegraphics[width=1.1in,height=1.25in,clip,keepaspectratio]{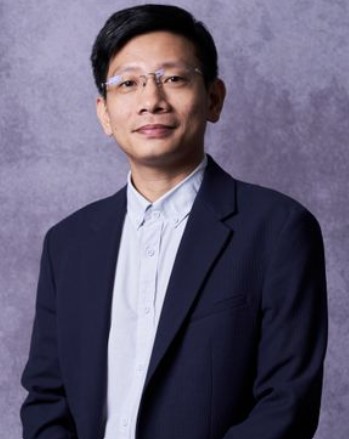}}]{Van-Dinh Nguyen}(Senior Member, IEEE) has been an Assistant Professor with VinUniversity, Vietnam, since September 2022. He was a Research Associate with SnT-University of Luxembourg, a Post-Doctoral Researcher and a Lecturer with Soongsil University, a Post-Doctoral Visiting Scholar with the University of Technology Sydney, and a Ph.D. Visiting Scholar with Queen’s University Belfast. He has authored or co-authored over 90 papers published in international journals and conference proceedings. His current research activity is focused on Open RAN, wireless sensing, edge/fog computing, and AI/ML solutions for wireless communications. He received four Best Conference Paper awards and four Exemplary Editor Awards from IEEE COMMUNICATIONS LETTERS and IEEE OPEN JOURNAL OF THE COMMUNICATIONS SOCIETY. He has served as a reviewer for many top-tier international journals on wireless communications and as a technical program committee member for several flagship international conferences in related fields. He is an Editor of the IEEE OPEN JOURNAL OF THE COMMUNICATIONS SOCIETY and IEEE SYSTEMS JOURNAL and a Senior Editor of IEEE COMMUNICATIONS LETTERS.

\end{IEEEbiography}

\begin{IEEEbiography}
    [{\includegraphics[width=1.1in,height=1.25in,clip,keepaspectratio]{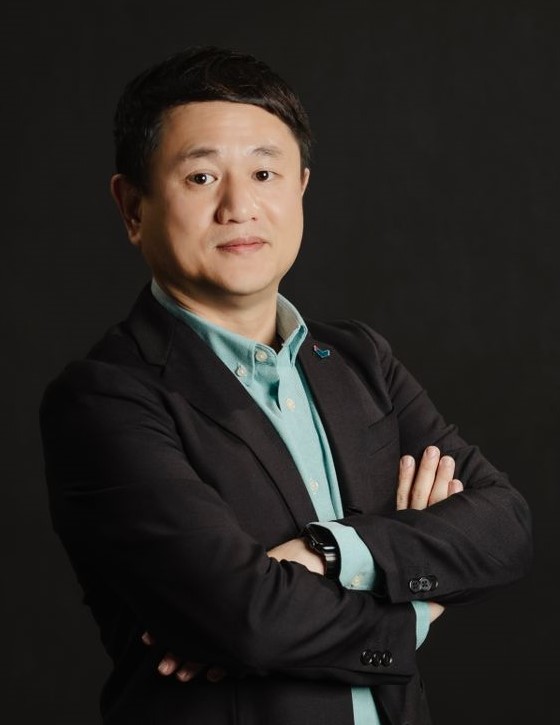}}]{Kok-Seng Wong}(Member, IEEE) has been an Associate Professor with the College of Engineering and Computer Science, VinUniversity, Vietnam, since 2020. He previously served as an Assistant/Associate Professor at Soongsil University, South Korea, an Assistant Professor at Nazarbayev University, Kazakhstan, and a Visiting Scholar at Beijing University of Posts and Telecommunications, China. His research interests include federated learning, privacy-preserving machine learning, trustworthy AI, and information security, with a focus on improving the efficiency, scalability, and practical deployment of AI systems. He has authored or co-authored over 80 papers in reputable international journals and top-tier conferences, including TETC, TNSM, ACL, CVPR, ECCV, ICLR, NeurIPS, and TheWebConf. He leads the Security and AI Lab (SAIL) and has served as a reviewer for leading journals and conferences on AI, machine learning, and security. 

\end{IEEEbiography}

\begin{IEEEbiography}[{\includegraphics[width=1in,height=1.25in,clip,keepaspectratio]{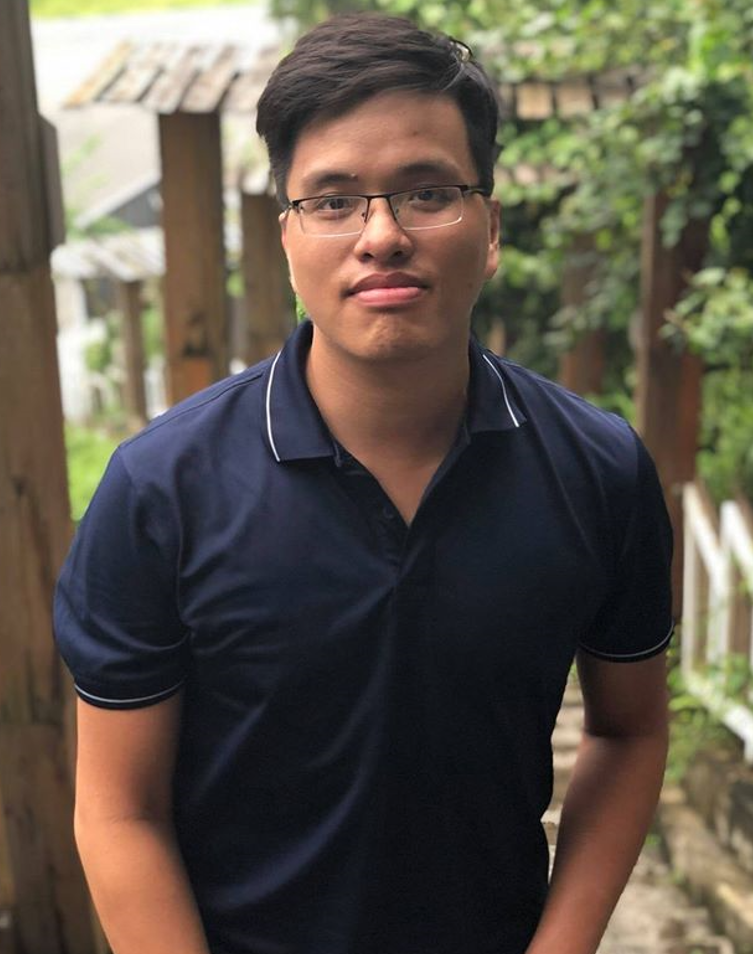}}]{Minh-Duong Nguyen} 
received a B.S. degree in electronics and telecommunications engineering from the Hanoi University of Science and Technology, Vietnam, in 2016. He was a DSP engineer in Viettel R\&D Center from 2017 to 2019. He was working in electronic warfare (i.e., electronic warfare systems and GPS/4G jamming systems).  
He was a senior embedded engineer in Vinsmart, Vingroup, from 2019 to 2020, developing physical layers for the 5G Base Station. He achieved his Ph.D.'s degree in 2025 at Pusan National University, South Korea. He is currently postdoctoral researcher in VinUniversity, Vietnam.

His research interests include reinforcement learning, federated learning, multi-task learning, meta-learning, domain generalization, data representation, and semantic communication. 
\end{IEEEbiography}

\begin{IEEEbiography}
[{\includegraphics[width=1.1in,height=1.25in,clip,keepaspectratio]{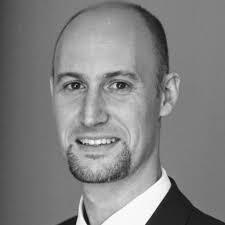}}]{SYMEON CHATZINOTAS} (Fellow, IEEE) received the M.Eng. degree in telecommunications from Aristotle University of Thessaloniki, Greece, in 2003, and the M.Sc. and Ph.D. degrees in electronic engineering from the University of Surrey, U.K., in 2006 and 2009, respectively. He is currently a Full Professor/Chief Scientist I and the Head of the Research Group SIGCOM, Interdisciplinary Centre for Security, Reliability and Trust, University of Luxembourg. In the past, he has lectured as a Visiting Professor with the University of Parma, Italy, and contributed in numerous research and development projects for the Institute of Informatics and Telecommunications, National Center for Scientific Research “Demokritos,” the Institute of Telematics and Informatics, Center of Research and Technology Hellas, and Mobile Communications Research Group, Center of Communication Systems Research, University of Surrey. He has authored more than 700 technical papers in refereed international journals, conferences, and scientific books. Prof. Chatzinotas received numerous awards and recognitions, including the IEEE Fellowship and an IEEE Distinguished Contributions Award. He is currently in the editorial board of the IEEE TRANSACTIONS ON COMMUNICATIONS, IEEE OPEN JOURNAL OF VEHICULAR TECHNOLOGY, and the International Journal of Satellite Communications and Networking.
\end{IEEEbiography}

\begin{IEEEbiography}
    [{\includegraphics[width=1.1in,height=1.25in,clip,keepaspectratio]{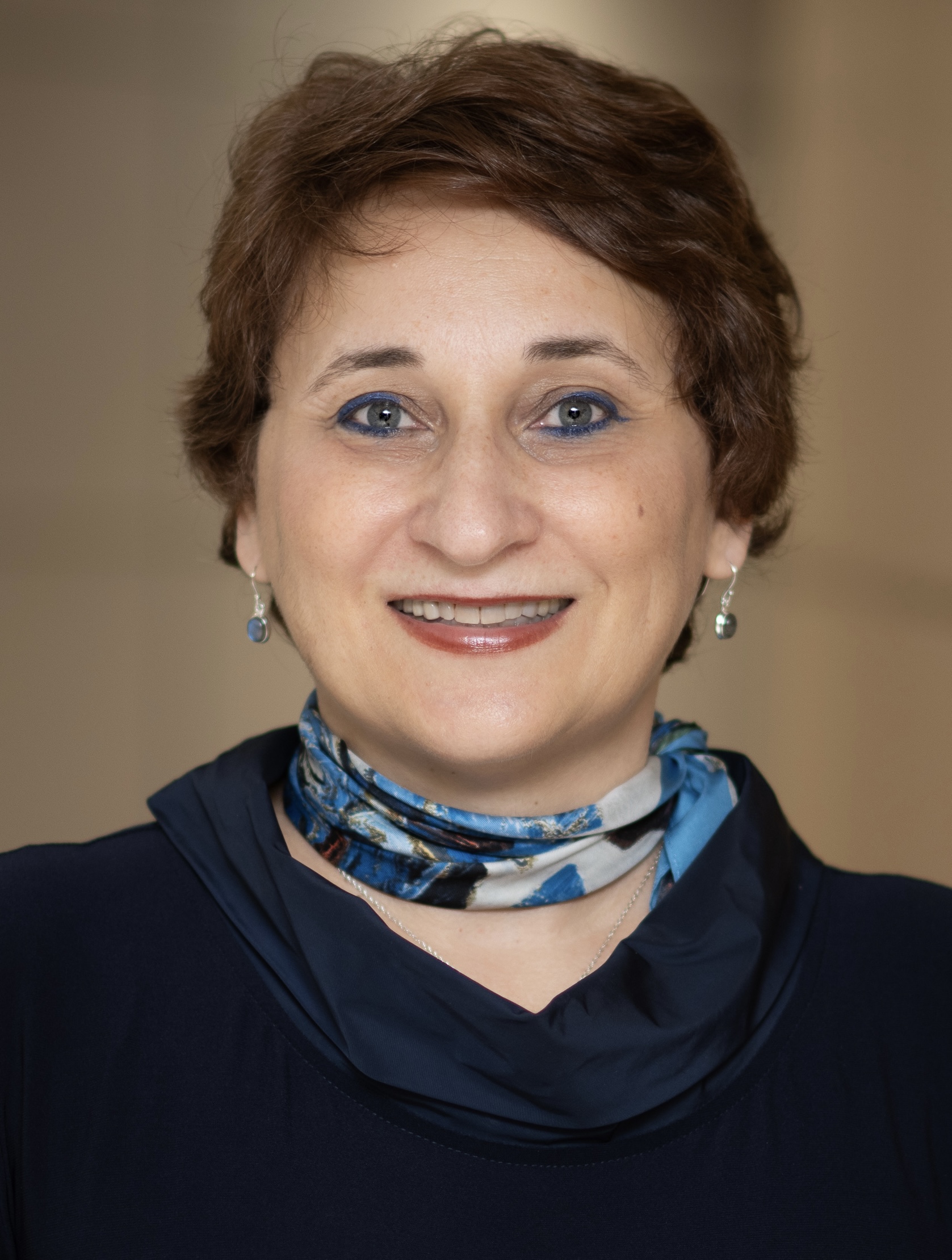}}]{Octavia A. Dobre }(Fellow, IEEE) 
    is a Professor and Tier-1 Canada Research Chair with Memorial University, Canada. Her research spans wireless communications and networking technologies, as well as optical and underwater communications. She has authored or co-authored over 600 publications in these areas and received ten Best Paper Awards, including the IEEE Communications Society Heinrich Hertz Award. 
    
    \noindent Dr. Dobre serves as the VP Publications of the \textit{IEEE Communications Society}. She was the founding Editor-in-Chief of the IEEE Open Journal of the Communications Society and previously served as Editor-in-Chief of \textit{IEEE Communications Letters}.
    
    \noindent Dr. Dobre is a Clarivate Highly Cited Researcher, a Fellow of the Royal Society of Canada, the Canadian Academy of Engineering, and the Engineering Institute of Canada. She is also an elected member of the European Academy of Sciences and Arts and the Academia Europaea.
\end{IEEEbiography}

\end{document}